\documentclass{article} 
\usepackage{iclr2027_conference,times}
\iclrfinalcopy

\usepackage{amsmath,amsfonts,bm}

\def\eqref#1{equation~\ref{#1}}

\def\1{\bm{1}}

\DeclareMathAlphabet{\mathsfit}{\encodingdefault}{\sfdefault}{m}{sl}
\SetMathAlphabet{\mathsfit}{bold}{\encodingdefault}{\sfdefault}{bx}{n}

\usepackage{subcaption}
\usepackage{wrapfig}
\usepackage{makecell}
\usepackage{booktabs}
\usepackage{caption}
\usepackage{multirow}
\usepackage{wrapfig}

\usepackage{tabularx}
\usepackage{array}

\usepackage{soul}
\usepackage{hyperref}
\usepackage{graphicx}
\usepackage{booktabs}
\usepackage{float}
\usepackage{overpic}
\usepackage{float}  
\usepackage{subfloat}
\usepackage{stfloats} 
\usepackage[skip=0.5\baselineskip]{caption}
\usepackage{bm}
\usepackage{amsmath}
\usepackage{amssymb}
\usepackage{booktabs}
\usepackage{multirow}
\usepackage{multicol}
\usepackage{enumitem}
\usepackage{subcaption}
\usepackage{threeparttable}
\usepackage{xspace}
\usepackage{color}
\usepackage[normalem]{ulem}
\usepackage{algorithmicx,algorithm}
\usepackage{algpseudocode}
\usepackage{algorithm,algpseudocode}
\usepackage{marvosym}   
\usepackage{colortbl}   
\usepackage{tcolorbox}  
\tcbuselibrary{breakable}   
\usepackage{wasysym}    
\usepackage[figuresright]{rotating} 

\usepackage{xspace}

\begin{document}
\begin{sloppypar}   
\title{PlanPath: Exploration-Guided Orchestration\\ for Patient-Level Pathology Reasoning}



\author{
\begin{tabular}{@{}c@{}}
Jiusong Ge$^{1}$, Mingzhao He$^{1}$, Qidong Liu$^{1,*}$, Chunbao Wang$^{1}$, Chunze Yang$^{1}$, Di Zhang$^{1}$ \\
Jiashuai Liu$^{1}$, Mireia Crispin-Ortuzar$^{2}$, Huazhu Fu$^{3}$, Chen Li$^{4,1,*}$, Zeyu Gao$^{2,*}$ \\[0.6ex]
{\normalfont\small $^{1}$Xi'an Jiaotong University \quad $^{2}$University of Cambridge \quad $^{3}$A*STAR} \\
{\normalfont\small $^{4}$Chinese Academy of Medical Sciences} \\
{\normalfont\small $^{*}$Corresponding author}
\end{tabular}
}






\maketitle

\begin{abstract}

A patient's pathology case spans tens of slides from distinct specimens and anatomical sites, each carrying non-redundant, site-specific evidence. Forming a patient-level conclusion therefore requires locating the slides relevant to a given question and integrating complementary findings across them.
However, existing pathology agents analyze one slide at a time and describe morphology rather than measure it, leaving evidence acquisition across a patient's specimens and patient-level quantification underexplored.
To address this gap, we propose \textbf{PlanPath}, an agentic framework for patient-level pathology visual question answering~(VQA). 
PlanPath turns the question into an exploration plan that specifies the evidence required, routes to the slides likely to carry it, and drives coarse-to-fine acquisition from tissue architecture down to individual cells.
Tissue- and cell-level tools supply the counts, proportions, and extents, which PlanPath aggregates across slides into a traceable answer.
To enable evaluation, we build two patient-level VQA datasets for renal and gastric cancer, averaging 15 WSIs per patient, with pathologist-verified question–answer pairs derived from diagnostic reports.
Experimental results show that PlanPath outperforms the strongest baseline by 9.4\% and 12.1\% in overall accuracy on the two datasets and further improves existing slide-level agents when plugged into our framework. The dataset and code are available at:
\url{https://github.com/JSGe-AI/PlanPath_1}.

\end{abstract}

\section{Introduction}

Recent advances in pathology foundation models and vision--language models have improved morphological representation and image--text alignment in computational pathology, providing reusable technical foundations for whole-slide image (WSI) understanding \citep{chen2024towards,lu2024visual}. Building upon these capabilities, pathology agents powered by large language models (LLMs) can autonomously localize relevant regions, invoke specialized analysis tools, and integrate diagnostic findings in response to clinical questions \citep{chen2025pathagent,ghezloo2025pathfinder}. 
These approaches have demonstrated promising performance on slide-level tasks while making the visual evidence behind their conclusions more explicit.

However, to better translate these advances into clinical diagnostic practice, pathology analysis should move beyond individual slides toward the patient level. Such analysis can integrate complementary evidence distributed across distinct specimens and anatomical sites, thereby providing a more complete and clinically representative characterization of a patient's pathological state. Nevertheless, this setting remains relatively underexplored in existing studies. To bridge this gap, we formally define \emph{patient-level pathology visual question answering} (VQA) as the task of generating a patient-level answer with explicit visual grounding from a patient's collection of slides and a diagnostic question through cross-slide, multiscale evidence acquisition.

\begin{figure}[t]
\begin{center}
\includegraphics[width=1.0\linewidth]{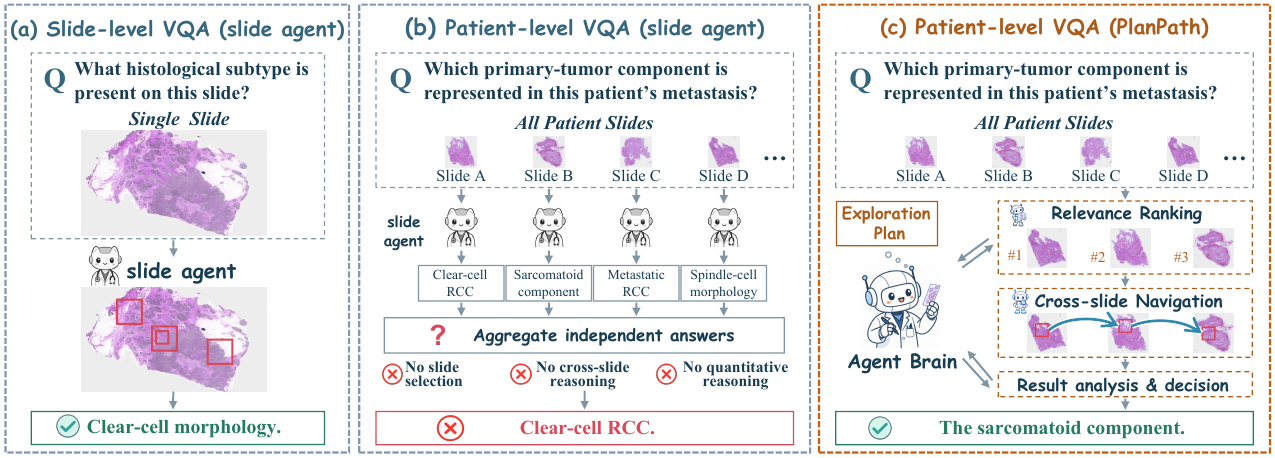}
\end{center}
\caption{Comparison of slide-level and patient-level pathology VQA. (a) A slide-level agent performs well on single-slide questions. (b) Directly applying slide-level agents to multiple slides lacks effective slide selection and cross-slide reasoning. (c) PlanPath addresses patient-level questions through exploration planning, slide ranking, and cross-slide evidence integration.}
\vspace{-3mm}
\label{fig1}
\end{figure}

Despite their effectiveness in slide-level analysis, existing pathology agents \citep{sun2026cpathagent,lyu2025wsi} remain limited for patient-level tasks. 
A naive extension is to aggregate answers independently generated from individual slides.
However, such aggregation has a fundamental limitation: it neither selects question-relevant slides nor integrates complementary evidence across them (Fig.~\ref{fig1}b).
This is critical because different patient-level questions often have different evidence requirements, while the relevant evidence is unevenly distributed across a patient's slide collection \citep{slideselection,huang2026act}. \textit{Effective patient-level analysis therefore requires first identifying slides relevant to the given question and then integrating complementary evidence across them}, rather than treating individual slides as independent units of analysis (Fig.~\ref{fig1}c).

Beyond slide selection, patient-level diagnosis often relies not only on the presence of pathological findings, but also on their quantitative characterization, including their amount, proportion, extent, and distribution. For example, the Gleason score of prostate cancer is determined by the relative proportions of its predominant and secondary patterns \citep{van20202019}. 
Existing pathology agents \citep{sun2026cpathagent,pathnavigate} reason over a limited set of local observations; while they can recognize local morphology, they struggle to translate it into reliable patient-level quantitative conclusions.
\textit{Effective patient-level analysis therefore also requires quantifying findings and aggregating them across slides at appropriate spatial scales.}

To address these challenges, we propose \textbf{PlanPath}, a pathology agentic framework for patient-level visual question answering. Given a patient-level question, PlanPath formulates an exploration plan that specifies the required diagnostic evidence and guides question-driven slide selection and cross-slide exploration. It then performs hierarchical cross-slide evidence acquisition, characterizing question-relevant regions through morphological observations, while tissue-level analysis provides spatial context and cell-level quantification is invoked when quantitative evidence is required. These complementary findings are integrated across slides at the patient level, with tissue- and cell-level analyses supporting measurements of areas, counts, proportions, and cross-slide distributions for questions that demand quantitative reasoning. Notably, as a general patient-level orchestration framework, PlanPath can adapt existing slide-level pathology agents to patient-level tasks.

Moreover, to support systematic evaluation, we collaborate with pathologists to construct two patient-level pathology VQA datasets for renal and gastric cancer. Questions and reference answers are derived from original diagnostic records and available slide evidence, with explicit verification that the anatomical sites required to answer each question are supported by corresponding slides. 

The main contributions of this work are summarized as follows:
\begin{itemize}[leftmargin=*]
    \item We first formally define \emph{patient-level pathology VQA} as a distinct task that requires reasoning over a patient's collection of slides rather than individual slides. To tackle this task, we propose \textbf{PlanPath}, a patient-level pathology agentic framework for question-driven slide selection, cross-slide evidence acquisition, and complementary evidence integration.
    
    \item We develop a multiscale evidence acquisition and quantitative analysis strategy that integrates slide-level context, tissue-level structure, and cell-level measurements, enabling aggregation of evidence for patient-level quantitative reasoning beyond isolated local observations.

    \item We construct two patient-level pathology VQA datasets for renal and gastric cancer, comprising 3,245 WSIs from 213 patients and 1,623 pathologist-verified question--answer pairs, and use them as standardized benchmarks to systematically evaluate PlanPath and a broad range of existing methods under the patient-level setting.
\end{itemize}


\section{Dataset Construction and Problem Formulation}
\label{sec:dataset}

This section presents the construction and quality control procedures for two patient-level pathology visual question answering (PatientVQA) datasets, analyzes their scale, question types, and diagnostic content, and formally defines the task inputs, outputs, and objectives. 

\subsection{Patient-Level Pathology VQA Dataset Construction}
\label{sec:dataset_construction}

We construct PatientVQA-Gastric and PatientVQA-Renal using real-world pathology records from patients with renal and gastric cancer. We include patients with definitive diagnoses, collect the corresponding pathology reports, and organize WSIs from different sampling sites into patient-level slide collections. GPT-5.6 Sol is used to derive candidate question--answer pairs from the reports, with reference answers grounded in explicitly documented diagnostic findings. The questions cover tumor type, grade, local invasion, margin status, lymph node involvement, and pathological staging. For quality control, two pathologists cross-review the candidate pairs and exclude those with medical or logical errors, ambiguous wording, insufficient support from the reports, or inadequate visual evidence in the available slides. The retained pairs are linked to their corresponding patient-level slide collections to form the final datasets. 

\subsection{Dataset Statistics and Analysis}
\label{sec:dataset_statistics}

The two datasets comprise 213 patients, 3,245 WSIs, and 1,623 question--answer pairs. PatientVQA-Gastric contains 104 patients, 2,227 WSIs, and 923 question--answer pairs, while PatientVQA-Renal contains 109 patients, 1,018 WSIs, and 700 question--answer pairs. The average numbers of slides per patient are 21.41 and 9.34, with ranges of 13--58 and 2--36, respectively, enabling evaluation across patient-level slide collections of varying sizes.

Both datasets include yes/no, single-answer multiple-choice (MCQ), and short-answer (Open) questions, accounting for 29.0\%, 28.1\%, and 42.9\% of PatientVQA-Gastric, and 46.7\%, 34.6\%, and 18.7\% of PatientVQA-Renal, respectively. These question types cover the presence of pathological findings, predefined diagnostic categories, and diagnostic answers. The diagnostic content covers tumor type, grade, local invasion, margin status, and pathological staging. Each patient is associated with multiple questions with varying slides and evidence requirements, supporting evaluation of morphological recognition, question-driven slide selection, and cross-slide evidence integration.

\subsection{Problem Formulation of Patient-level VQA}
\label{sec:problem_formulation}

Let $\mathcal{S}_p = \{W_{p,i}\}_{i=1}^{N_p}$ denote the WSI collection of patient $p$. Given a diagnostic question $q$, patient-level pathology VQA aims to acquire evidence from $\mathcal{S}_p$ and generate a visually grounded answer:
\begin{equation}
    (\hat{a}, \mathcal{E})
    = \mathcal{F}(\mathcal{S}_p, q),
\end{equation}
where $\mathcal{F}$ is the answering system, $\hat{a}$ the predicted answer, and $\mathcal{E}$ the supporting evidence. 

The evidence set $\mathcal{E}$ contains question-relevant visual observations and, when needed, quantitative measurements, each linked to its source slide, spatial region, and observation scale. Depending on the question, evidence may be acquired from one or multiple WSIs.

The task therefore requires selecting relevant slides and regions, acquiring evidence at appropriate scales, and integrating it into an accurate, evidence-grounded patient-level answer.
\section{Method}

\begin{figure}[t]
\begin{center}
\includegraphics[width=1.0\linewidth]{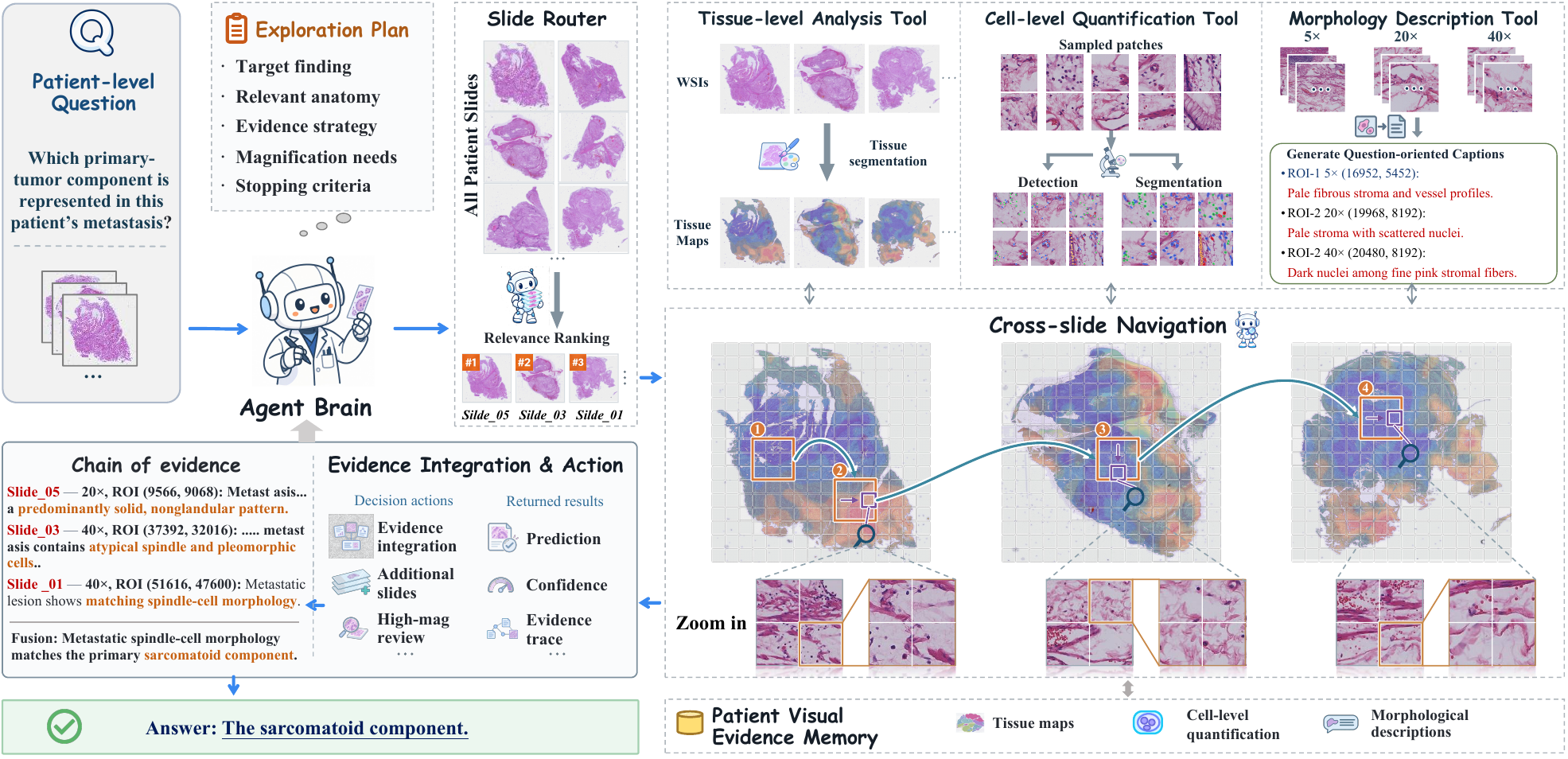}
\end{center}
\caption{Overview of PlanPath. The Agent Brain plans and coordinates question-driven slide retrieval and cross-slide multiscale evidence acquisition, followed by iterative evidence integration for prediction, confidence estimation, and traceable reasoning.}
\vspace{-3mm}
\label{planpath}
\end{figure}

\subsection{Framework Overview}

We propose \textbf{PlanPath}, an agentic framework for patient-level pathology VQA. As illustrated in Fig.~\ref{planpath}, a unified \emph{Agent Brain} coordinates the \emph{Slide Router}, cross-slide navigation, and analysis tools: the \emph{Tissue-level Analysis Tool}, \emph{Cell-level Quantification Tool}, and \emph{Image Description Tool}. 

The Agent Brain first constructs an exploration plan that specifies the required evidence, analysis scales, and stopping criteria, denoted as \textbf{Question-driven Exploration Planning}. Guided by this plan, the Slide Router retrieves relevant slides through \textbf{Patient-level Slide Routing}. PlanPath then performs \textbf{Hierarchical Evidence Acquisition} by progressively exploring candidate ROIs across slides and magnifications and invoking analysis tools as needed. All acquired evidence is stored with its source slide, ROI, magnification, and spatial lineage, enabling the patient-level \textbf{Evidence Integration and Answer Generation} to remain traceable to visual and quantitative evidence.





\subsection{Question-Driven Exploration Planning}

PlanPath first transforms the original question $q$ into a structured exploration plan that guides subsequent evidence acquisition:
\begin{equation}
\Pi(q)
=
\left(
\mathcal{D},
\tau,
\mathcal{T},
\mathcal{R},
\mathcal{G}
\right),
\label{eq:plan}
\end{equation}
where $\mathcal{D}$ denotes a set of semantic descriptions used for slide retrieval, $\tau$ represents the logical property of the question, $\mathcal{T}$ specifies the target evidence, $\mathcal{R}$ defines the required spatial resolutions and tool calls, and $\mathcal{G}$ specifies evidence sufficiency and stopping criteria.

According to their reasoning requirements, input questions are categorized into \emph{existential}, \emph{exhaustive}, \emph{extremum}, \emph{quantitative}, or composite types. These categories determine the required evidence coverage. For example, an existential question may terminate once reliable direct positive evidence is obtained, whereas patient-level negative conclusions, extrema, extent-related assessments, and other global judgments generally require broader slide coverage. The derived exploration plan also serves as a tool-routing mechanism. 
Based on $\Pi$, the Agent Brain determines which analysis tools are required for the current question and specifies their roles in evidence acquisition. Tools irrelevant to the current question are excluded from the execution path. In conclusion, PlanPath does not apply a fixed analysis pipeline to every slide. Instead, it first determines \emph{what evidence is required} and decides \emph{where, at what scale, and with which tools} to acquire it.

\subsection{Patient-Level Slide Routing}

The Slide Router identifies and prioritizes slides that are likely to contain evidence relevant to the current question. 
Given the patient WSI collection $\mathcal{S}$ and the retrieval descriptions $D$ specified in the exploration plan $\Pi$, it produces a relevance-ranked candidate slide pool for evidence acquisition. The Slide Router is used only to determine the order and scope of exploration. 
Routing scores are not considered diagnostic evidence and do not directly contribute to the final answer. 
Any retrieved slide must undergo ROI-level analysis before it can support a patient-level conclusion.

\textbf{Candidate Pool Initilization.} For the patient WSI collection $\mathcal{S}$, the Slide Router evaluates the relevance of each slide according to the retrieval descriptions $\mathcal{D}$ defined in the exploration plan. For slide $S_i$, its routing score is computed as:
\begin{equation}
r_i
=
\max_{d \in \mathcal{D}}
\operatorname{sim}
\left(
f_{\mathrm{slide}}(S_i),
f_{\mathrm{text}}(d)
\right),
\label{eq:slide_routing}
\end{equation}
where $f_{\mathrm{slide}}(\cdot)$ and $f_{\mathrm{text}}(\cdot)$ are the slide/text encoders, and $\operatorname{sim}(\cdot,\cdot)$ denotes cross-modal similarity.

Then, the slides are ranked according to $r_i$, and an initial candidate pool is constructed as:
\begin{equation}
\mathcal{C}_0
=
\operatorname{TopK}_{S_i \in \mathcal{S}}
\left(
r_i
\right).
\label{eq:candidate_pool}
\end{equation}
Using multiple complementary retrieval descriptions allows slides containing different but relevant morphological patterns to enter the candidate pool.

\textbf{Candidate Pool Extension.} Importantly, $\mathcal{C}_0$ is an \emph{initial} candidate pool rather than a fixed retrieval set. If all slides in the current pool have been explored but the required patient-level evidence remains insufficient, the unmet evidence requirements are fed back to the Slide Router. The router then retrieves additional slides from the remaining WSI collection:
\begin{equation}
\mathcal{C}_{t+1}
=
\mathcal{C}_{t}
\cup
\operatorname{TopK}
\left(
\mathcal{S}
\setminus
\mathcal{C}_{t}
\right),
\label{eq:candidate_expansion}
\end{equation}
where retrieval is conditioned on the updated evidence requirements. The retrieved slides are appended to the candidate pool and undergo the hierarchical evidence acquisition procedure.

Notably, the extent of candidate coverage is also determined by the question type $\tau$. For focal existential findings, PlanPath prioritizes high-ranking slides and may terminate once sufficient positive evidence is obtained. For negative, exhaustive, extremum, or extent-related tasks, broader coverage is maintained to reduce the risk of missing critical evidence that appears only on a subset of slides.

\subsection{Hierarchical Evidence Acquisition}

For each candidate slide, PlanPath adopts a coarse-to-fine strategy to progressively localize, inspect, and quantify question-relevant pathological evidence across multiple spatial scales. Specifically, evidence acquisition is organized around three complementary components: tissue context, morphological observation, and cell-level quantification.

\textbf{Coarse-to-Fine ROI Selection.}
PlanPath first establishes global tissue context from low-magnification views and identifies candidate ROIs according to the exploration plan. Each candidate ROI $R$ is prioritized based on semantic relevance, tissue context, and image quality:
\begin{equation}
s(R)
=
\alpha s_{\mathrm{sem}}(R)
+
\beta s_{\mathrm{tissue}}(R)
+
\gamma s_{\mathrm{quality}}(R),
\label{eq:roi_score}
\end{equation}
where the three terms measure question-related semantic relevance, consistency with target tissue structures, and image quality, respectively. High-priority ROIs are progressively examined at higher magnifications while preserving their parent--child spatial relationships.

\textbf{Tissue-Level Analysis.}
When tissue composition, lesion extent, or spatial interfaces are relevant to the question, the Tissue-level Analysis Tool extracts tissue regions, areas, distributions, and interfaces from the WSI. These outputs provide global spatial context and guide subsequent ROI selection rather than directly determining the diagnosis. For quantitative tasks, tissue-level measurements can further support the estimation of areas, proportions, and spatial distributions.

\textbf{Morphological Observation.}
For selected ROIs, the Image Description Tool generates morphological observations at the corresponding magnification. Low-magnification views characterize tissue architecture and spatial organization, whereas higher-magnification views capture cellular arrangement and fine-grained morphology. To separate visual observation from diagnostic reasoning, the tool describes the visible pathological findings rather than directly predicting the final answer.

\textbf{Cell-Level Quantification.}
When quantitative evidence is required, the Cell-level Quantification Tool analyzes selected high-magnification ROIs to obtain cell counts, proportions, densities, and spatial statistics. Quantification is restricted to ROIs satisfying relevance, image-quality, and tissue-coverage requirements, with each measurement linked to its source slide and ROI.

\textbf{Evidence Recording and Iterative Exploration.}
Each acquired observation is stored as a structured evidence record:
\begin{equation}
e_j
=
\left(
S_j,
R_j,
m_j,
v_j,
z_j,
\ell_j
\right),
\label{eq:evidence_record}
\end{equation}
where $S_j$, $R_j$, $m_j$, $v_j$, $z_j$, and $\ell_j$ denote the source slide, ROI, magnification, morphological observation, optional quantitative measurement, and spatial lineage, respectively.

After each step, the Agent Brain evaluates evidence and selects the next action:
\begin{equation}
a_t
=
\pi\left(
q,
\Pi,
\mathcal{E}_t,
\mathcal{C}_t
\right),
\qquad
a_t
\in
\left\{
\texttt{stop},
\texttt{zoom},
\texttt{next-roi},
\texttt{next-slide},
\texttt{reroute}
\right\}.
\label{eq:agent_action}
\end{equation}
Here, $\mathcal{E}_t$ denotes evidence accumulated by step $t$, and $\mathcal{C}_t$ the candidate slide pool. These actions determine whether PlanPath stops, zooms into the ROI, moves to another ROI or slide, or returns to the Slide Router for additional slides when the pool is insufficient. This process allows evidence acquisition to adapt to observed evidence rather than follow a fixed traversal of all slides and ROIs.

\textbf{Cross-Scale Quantitative Aggregation.}
For questions requiring patient-level quantitative reasoning, valid measurements obtained from tissue- and cell-level analyses are aggregated according to the quantitative requirement of the question. Let $z_i$ denote the valid measurements obtained from slide $S_i$. Patient-level quantitative evidence is computed as:
\begin{equation}
Z_p
=
\operatorname{Agg}_{\tau}
\left(
\{z_i\}_{i \in \mathcal{C}_{\mathrm{cov}}}
\right),
\label{eq:quantitative_aggregation}
\end{equation}
where $\mathcal{C}_{\mathrm{cov}}$ denotes the set of slides with valid measurements. Depending on the task, the aggregation may correspond to counting, proportion estimation, extremum selection, or preservation of cross-slide distributions. Overlapping ROIs are excluded or corrected to avoid double counting, and numerical conclusions are reported only when predefined coverage and quality requirements are satisfied. Through this hierarchical process, PlanPath links tissue context, regional morphology, and cell-level measurements into a unified and traceable evidence chain for patient-level reasoning.

\subsection{Evidence Integration and Answer Generation}

All visual observations and quantitative measurements are stored in a patient-level evidence repository $\mathcal{E}_t=\{e_j\}$. Rather than aggregating independent slide-level predictions, PlanPath integrates complementary cross-slide evidence according to the exploration plan. Before termination, an Evidence Gate evaluates whether the accumulated evidence satisfies the requirements defined by $\Pi$:
\begin{equation}
G\left(
\mathcal{E}_t \mid \Pi
\right)
=
\mathbf{1}
\left\{
C_t \geq C_{\min}
\land
Q_t \geq Q_{\min}
\land
T_t = 1
\land
L_t = 1
\right\},
\label{eq:evidence_gate}
\end{equation}
where $C_t$, $Q_t$, $T_t$, and $L_t$ denote coverage, quality, target satisfaction, and spatial consistency.

If the Evidence Gate is not satisfied, PlanPath continues the iterative exploration in Sec.~3.4 according to the remaining evidence requirements until sufficient evidence is obtained or the available slides or analysis budget are exhausted. Once satisfied, the Agent Brain constructs a compact evidence context $\mathcal{E}^{*}$ and generates the final answer:
\begin{equation}
\hat{y}
=
\operatorname{Dec}
\left(
q,
\Pi,
\mathcal{E}^{*}
\right).
\label{eq:answer_generation}
\end{equation}
The output follows the original question format and retains confidence, evidence sufficiency, and traceable evidence identifiers. If exploration ends before the requirements are fully satisfied, the answer is marked as low-confidence and evidence-insufficient.

\section{Experiment}

\begin{table*}[t]
\centering
\caption{
Performance comparison on PatientVQA-Gastric and PatientVQA-Renal.
The best result in each column is shown in \textbf{bold}, and the second-best result is \underline{underlined}.
}
\label{tab:main_results}

\setlength{\tabcolsep}{5pt}
\renewcommand{\arraystretch}{1.08}

\resizebox{\textwidth}{!}{
\begin{tabular}{lccccccccc}
\toprule
\multirow{2}{*}{\textbf{Method}}
& \multirow{2}{*}{\textbf{Input}}
& \multicolumn{4}{c}{\textbf{PatientVQA-Gastric}}
& \multicolumn{4}{c}{\textbf{PatientVQA-Renal}} \\
\cmidrule(lr){3-6}
\cmidrule(lr){7-10}

&
& \textbf{Total}
& \textbf{MCQ}
& \textbf{Yes/No}
& \textbf{Open}
& \textbf{Total}
& \textbf{MCQ}
& \textbf{Yes/No}
& \textbf{Open} \\
\midrule

\multicolumn{10}{l}{
\footnotesize\color{gray!70}\textit{Pretrained MLLM Models}
} \\

GPT-5.5
& Slide-T
& 35.10
& 39.77
& 30.97
& 34.85
& 40.71
& 23.97
& 60.55
& 22.14 \\

GPT-5.6-Luna
& Slide-T
& 37.70
& 36.29
& 50.75
& 29.80
& 49.29
& 47.11
& 62.08
& 21.37 \\

Qwen3.5-4B
& Slide-T
& 50.38
& 42.86
& 65.30
& \underline{45.20}
& 55.14
& 50.00
& 69.11
& 29.77 \\

Qwen3.5-9B
& Slide-T
& 47.24
& 41.70
& 62.31
& 40.66
& 55.57
& 46.69
& 71.56
& \underline{32.06} \\

\midrule
\multicolumn{10}{l}{
\footnotesize\color{gray!70}\textit{Supervised Pathology Models}
} \\
MedDr
& Slide-T
& 30.44
& 30.50
& 64.18
& 7.58
& 55.86
& 58.26
& 70.34
& 15.27 \\

WSI-VQA
& WSI
& 32.83
& 44.79
& 47.39
& 15.15
& 15.86
& 13.22
& 20.18
& 9.92 \\

TITAN
& WSI
& --
& 26.64
& 39.93
& --
& --
& 13.64
& 57.80
& -- \\

Quilt-LLaVA
& Patch
& 29.47
& 51.35
& 35.82
& 10.86
& 37.00
& \underline{61.98}
& 29.97
& 8.40 \\

WSI-LLaVA
& WSI
& 32.50
& 22.78
& \underline{67.91}
& 14.90
& 49.00
& 35.12
& 71.87
& 17.56 \\

MedGemma-1.5-4B
& Slide-T
& \underline{50.81}
& \underline{52.12}
& 64.93
& 40.40
& \underline{60.57}
& 59.50
& \underline{73.39}
& 30.53 \\

SlideChat
& WSI
& 34.78
& 26.25
& 63.43
& 20.96
& 44.43
& 17.77
& 71.56
& 25.95 \\

\midrule
\multicolumn{10}{l}{
\footnotesize\color{gray!70}\textit{Agentic Systems}
} \\

PathAgent
& WSI
& 33.91
& 40.93
& 38.81
& 26.01
& 33.14
& 46.28
& 30.28
& 16.03 \\

PathNavigate
& WSI
& 34.45
& 43.24
& 58.96
& 12.12
& 55.14
& 57.85
& 67.58
& 19.08 \\

GIANT
& WSI
& 37.27
& 44.40
& 33.21
& 35.35
& 47.14
& 34.30
& 63.30
& 30.53 \\

BEACON
& WSI
& --
& 33.98
& 38.81
& --
& --
& 47.11
& 25.08
& -- \\

\rowcolor{gray!10}
\textbf{PlanPath (ours)}
& WSI
& \textbf{60.24}
& \textbf{57.92}
& \textbf{70.15}
& \textbf{55.05}
& \textbf{72.71}
& \textbf{78.51}
& \textbf{74.92}
& \textbf{56.49} \\

\bottomrule
\end{tabular}
}

\vspace{-3mm}
\end{table*}


\subsection{Experimental Setup}
\textbf{Implementation Details.}
The \emph{Agent Brain} in PlanPath is instantiated with Qwen3.5-4B \citep{qwen2026qwen35}, while the \emph{Slide Router} is implemented using TITAN \citep{TITAN}. The \emph{Tissue-level Analysis Tool} employs a prototype-guided tissue segmentation model, the \emph{Cell-level Quantification Tool} uses HistoPLUS \citep{histoplus}, and the \emph{Image Description Tool} is based on Patho-R1-7B \citep{patho-r1}. Cross-slide navigation is jointly driven by the tissue-level analysis results and PLIP \citep{PLIP}: tissue maps provide spatial context, while PLIP evaluates the semantic relevance between candidate regions and the current evidence requirements. Together, these components support slide selection, multiscale evidence acquisition, and patient-level evidence integration. Details of each component are provided in \textbf{Appendix~\ref{app:implementation}}.

\textbf{Baselines.}
We compare PlanPath with three categories of methods: general-purpose multimodal Large Language Models, including GPT-5.5 \citep{openai2026gpt55}, GPT-5.6-Luna \citep{openai2026gpt56}, Qwen3.5-4B \citep{qwen2026qwen35}, and Qwen3.5-9B \citep{qwen2026qwen35}; \emph{Supervised Pathology Models}, including MedDr \citep{MedDr}, WSI-VQA \citep{wsi-vqa}, TITAN \citep{TITAN}, Quilt-LLaVA \citep{quilt-LLAVA}, WSI-LLaVA \citep{Wsi-llava}, MedGemma-1.5-4B \citep{medgemma}, and SlideChat \citep{slidechat}; and \emph{Agentic Systems}, including PathAgent \citep{chen2025pathagent}, PathNavigate \citep{pathnavigate}, GIANT \citep{giant}, and BEACON \citep{Beacon}. For slide-level methods, each slide is analyzed independently, after which Qwen3.5-4B aggregates the outputs from all slides into a patient-level answer. We evaluate these baselines using a majority-voting aggregation strategy, with results reported in \textbf{Appendix~\ref{app:aggregation}}.

\textbf{Evaluation Metrics.}
We use accuracy for MCQ, Yes/No, and open-ended short-answer questions, and report \emph{Total Accuracy} over all questions as the overall metric for patient-level VQA performance. BLEU-1 and ROUGE-L are reported in \textbf{Appendix~\ref{app:detailed_results}}.

\subsection{Main Results on Patient-Level Pathology VQA}

Table~\ref{tab:main_results} presents the main results on PatientVQA-Gastric and PatientVQA-Renal. PlanPath achieves the best overall performance on both benchmarks. On PVQA-Gastric, PlanPath reaches a Total Accuracy of 60.24\%, outperforming the strongest baseline by 9.43 percentage points. It also achieves the best performance on MCQ, Yes/No, and open-ended questions. The improvement is particularly notable for open-ended questions, where PlanPath achieves an accuracy of 55.05\%. On Renal, the performance improvement is even more pronounced. PlanPath achieves a Total Accuracy of 72.71\%, exceeding the strongest baseline by 12.14 percentage points. It reaches an MCQ accuracy of 78.51\% and an open-ended accuracy of 56.49\%. Overall, these results demonstrate that question-driven slide selection and cross-slide evidence integration are particularly effective for patient-level pathology VQA tasks that require integrating complementary information distributed across multiple WSIs.

\subsection{Compatibility with Slide-Level Agents}

\begin{figure}[t]
\centering

\begin{minipage}[t]{0.46\linewidth}
    \vspace{0pt}
    \centering
    \includegraphics[width=\linewidth]{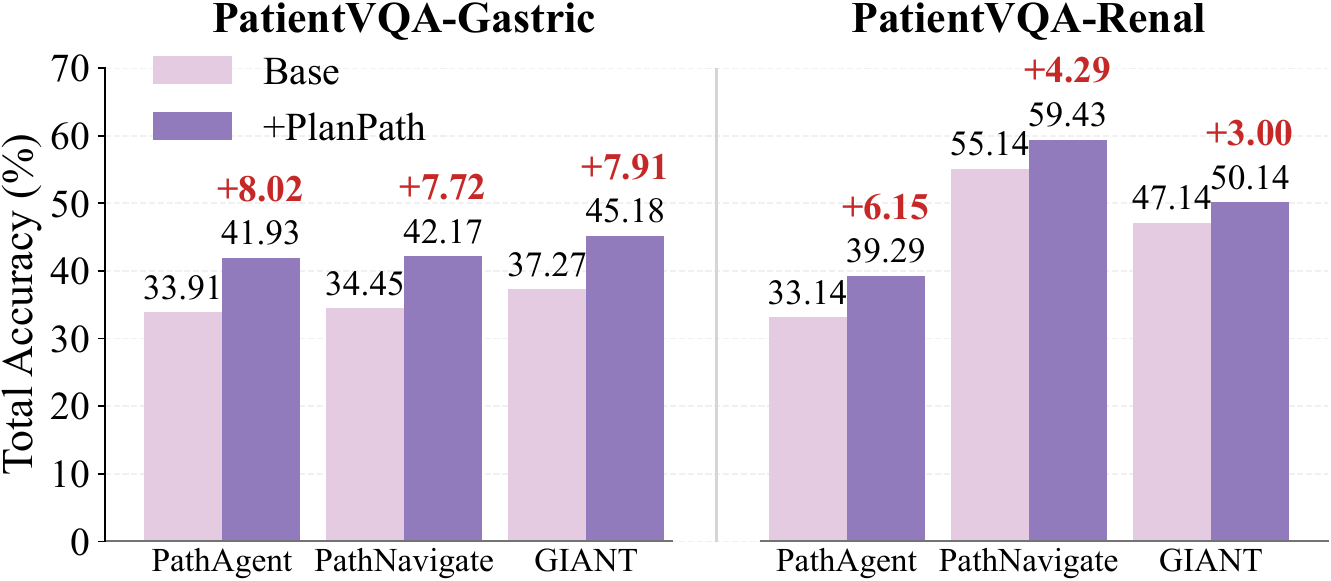}
    \caption{Compatibility of PlanPath with slide-level pathology agents.}
    \label{fig3}
\end{minipage}
\hfill
\begin{minipage}[t]{0.52\linewidth}
    \vspace{0pt}
    \centering

    \captionof{table}{
Evaluation of patient-level quantitative questions using exact-match and tolerance-based accuracy. All values are reported as \%.
}
    \label{tab2}

    \setlength{\tabcolsep}{3.5pt}
    \renewcommand{\arraystretch}{1.05}

    \resizebox{\linewidth}{!}{
    \begin{tabular}{lcccc}
    \toprule
    Method
    & PathAgent
    & PathNavigate
    & GIANT
    & \textbf{PlanPath}
    \\
    \midrule

    Exact Match
    & 5.56
    & 16.67
    & 11.11
    & \textbf{38.89}
    \\

    Tol. $\pm 5$
    & 5.56
    & 16.67
    & 16.67
    & \textbf{61.11}
    \\

    Tol. $\pm 10$
    & 11.11
    & 44.44
    & 38.89
    & \textbf{72.22}
    \\

    Tol. $\pm 15$
    & 11.11
    & 44.44
    & 38.89
    & \textbf{72.22}
    \\

    Tol. $\pm 20$
    & 38.89
    & 55.56
    & 66.67
    & \textbf{83.33}
    \\

    \bottomrule
    \end{tabular}
    }

\end{minipage}

\vspace{-1mm}
\end{figure}



We evaluate the compatibility of PlanPath with three representative slide-level pathology agents: PathAgent, PathNavigate, and GIANT. As shown in Fig.~\ref{fig3}, integrating PlanPath consistently improves all three agents on both patient-level datasets. On Gastric, the gains range from 7.72 to 8.02 percentage points, while improvements of 3.00--6.15 points are observed on Renal. These consistent gains demonstrate that PlanPath can effectively extend different slide-level agents to patient-level multi-slide reasoning without relying on a specific underlying agent. Detailed performance changes across individual question types are provided in \textbf{Appendix~\ref{app:compatibility}}.



\subsection{Quantitative Reasoning and Inference Efficiency}










We further evaluate PlanPath in terms of quantitative reasoning and inference efficiency. As shown in Table~\ref{tab2}, PlanPath consistently outperforms representative pathology agents under different quantitative evaluation criteria, achieving 38.89\% exact match and 61.11\%, 72.22\%, 72.22\%, and 83.33\% accuracy under tolerance thresholds of $\pm5$, $\pm10$, $\pm15$, and $\pm20$, respectively. 

Meanwhile, Table~\ref{tab3} compares prompt token consumption on the closed-ended question subset common to all methods, including BEACON, for fair comparison. On this cohort, PlanPath requires 164.32K and 120.03K tokens per question on Gastric and Renal, respectively, substantially fewer than PathAgent, PathNavigate, GIANT, and BEACON. These results indicate that question-driven slide routing and hierarchical evidence acquisition reduce unnecessary processing of irrelevant slides and redundant visual information while maintaining patient-level reasoning performance.

\subsection{Ablation and Case Study}

Table~\ref{tab4} evaluates key PlanPath components. Removing the Slide Router reduces accuracy from 65.62\% to 64.02\%, confirming the importance of question-driven slide selection. Removing the Tissue-level Analysis Tool causes the largest drop of 3.57 percentage points, highlighting the role of tissue context in relevant-region exploration. Removing Cell-level Quantification yields a smaller decrease to 65.19\%, suggesting a more task-dependent benefit. Removing both tissue- and cell-level tools further degrades performance, supporting the value of multiscale evidence acquisition.







\begin{figure}[t]
\begin{center}
\includegraphics[width=1.0\linewidth]{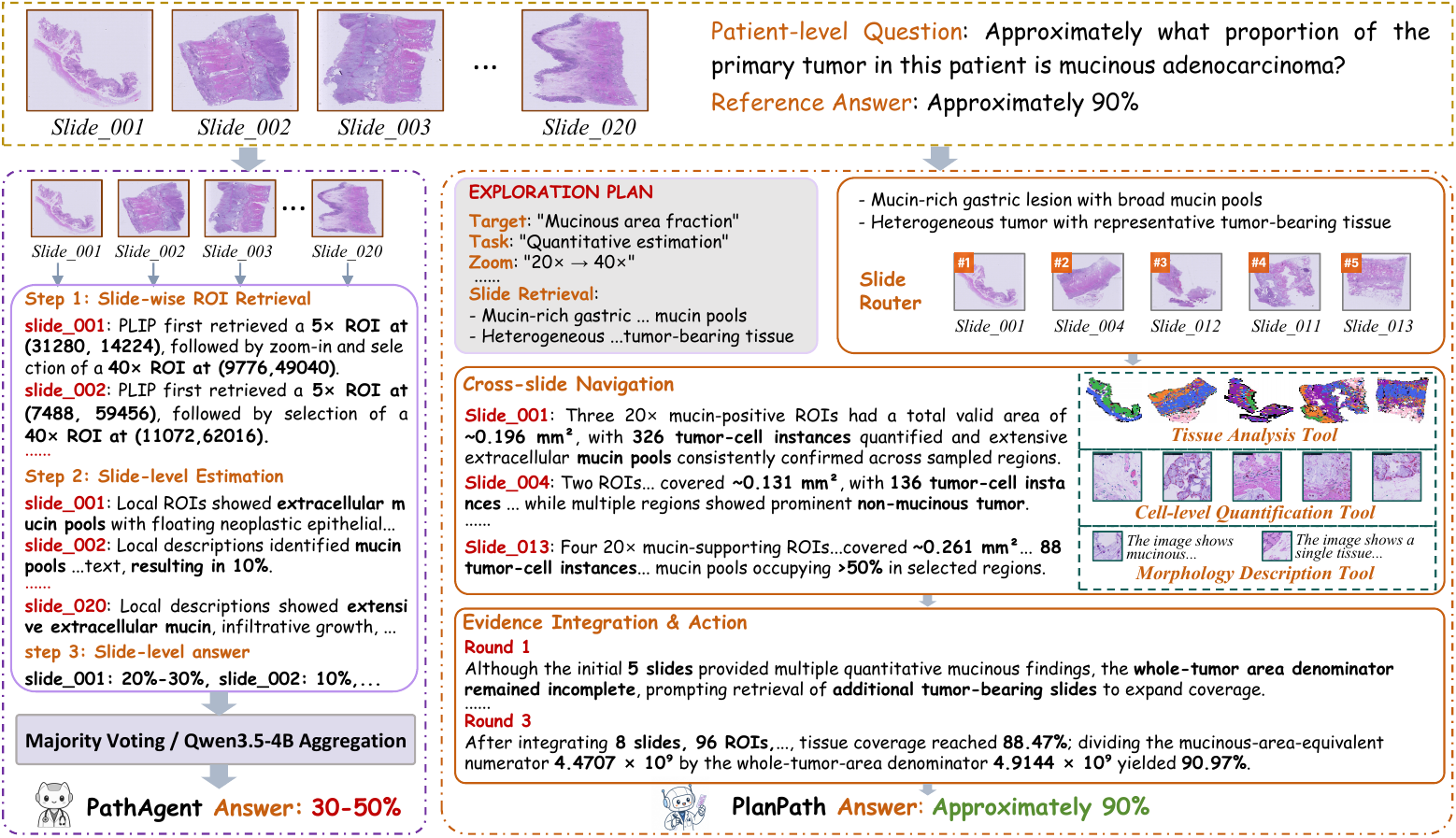}
\end{center}
\caption{Case study comparing slide-level agents and PlanPath for patient-level VQA.}
\vspace{-3mm}
\label{fig4}
\end{figure}

Figure~\ref{fig4} presents a quantitative VQA case. A slide-level agent analyzes WSIs independently and produces inconsistent estimates, leading to an inaccurate answer. PlanPath formulates an exploration plan, retrieves question-relevant slides, and performs cross-slide tissue-level, cell-level, and morphological analysis. When evidence is insufficient, it retrieves tumor-bearing slides. By integrating evidence across slides and scales, PlanPath estimates approximately 90\%, consistent with the reference answer, demonstrating the value of question-driven slide selection, multiscale analysis, and iterative cross-slide evidence integration. Additional Case studies are provided in \textbf{Appendix~\ref{app:qualitative}}.


\begin{table}[t]
\centering

\begin{minipage}{0.36\linewidth}
\centering
\small

\captionof{table}{
Prompt token consumption per patient-level question
(K tokens/Question \textdownarrow).
}
\label{tab3}

\resizebox{0.95\linewidth}{!}{
\begin{tabular}{lcc}
\toprule
Method & Gastric & Renal\\
\midrule
PathAgent & 874.42 & 499.50\\
PathNavigate & 234.69 & 128.90\\
GIANT & 1130.88 & 579.84\\
BEACON & 452.27 & 249.01 \\
\textbf{PlanPath} & \textbf{164.32} & \textbf{120.03}\\
\bottomrule
\end{tabular}
}

\end{minipage}
\hfill
\begin{minipage}{0.60\linewidth}
\centering
\small

\captionof{table}{
Ablation study of PlanPath components on two patient-level datasets.
Cell Quant. denotes the Cell-level Quantification Tool.
Results are reported as Total (\%).
}
\label{tab4}

\resizebox{\linewidth}{!}{
\begin{tabular}{lccc}
\toprule
Configuration & Renal & Gastric & Overall\\
\midrule
PlanPath (Full)
& \textbf{72.71}
& \textbf{60.24}
& \textbf{65.62}\\

w/o Slide Router
&71.29&58.50&64.02\\

w/o Tissue Analysis
&69.43&56.45&62.05\\

w/o Cell Quant.
&72.57&59.59&65.19\\

w/o Tissue Analysis \& Cell Quant.
&69.57&57.31&62.60\\

\bottomrule
\end{tabular}
}

\end{minipage}

\end{table}

\section{Related Works}


\textbf{Pathology Foundation and Vision--Language Models.} Recent pathology foundation models have improved transferable representation learning for histopathology. Vision foundation models such as Virchow \citep{vorontsov2024virchow}, Prov-GigaPath \citep{xu2024gigapath}, and CHIEF \citep{wang2024pathology} learn generalizable morphological representations through large-scale pretraining. Vision--language models such as PLIP \citep{PLIP} and CONCH \citep{CONCH} align pathological images with natural language. Approaches, including PathChat \citep{pathchat}, TITAN \citep{TITAN}, SlideChat \citep{slidechat}, and WSI-LLaVA \citep{Wsi-llava}, extend multimodal understanding and visual question answering to whole-slide images. These advances provide strong foundations for pathological representation learning and image--text reasoning, but most still operate on predefined image inputs or individual WSIs and do not explicitly address question-driven evidence acquisition and integration across multiple slides from the same patient.


\textbf{Pathology Agents for Whole-Slide Image Reasoning.} Recent pathology agents have further moved beyond static WSI prediction toward active diagnostic evidence acquisition through ROI localization, magnification adjustment, tool invocation, and iterative reasoning. Representative approaches such as PathAgent \citep{chen2025pathagent}, GIANT \citep{giant}, and PathNavigate \citep{pathnavigate} demonstrate the effectiveness of adaptive evidence acquisition for whole-slide image reasoning. However, most existing methods \citep{ghezloo2025pathfinder,lyu2025wsi,ge2026thinking} remain focused on within-slide exploration and primarily reason from local visual observations. They provide limited support for selecting question-relevant slides from a patient's WSI collection, dynamically expanding the slide search when evidence is insufficient, and systematically aggregating quantitative evidence across slides. PlanPath extends pathology agents from within-slide navigation to patient-level cross-slide evidence acquisition and integration.

\section{Conclusion}

In this work, we formally define \emph{patient-level pathology visual question answering} and propose \textbf{PlanPath}, a patient-level agentic framework for question-driven slide selection, hierarchical evidence acquisition, quantitative analysis, and cross-slide integration. We also construct two patient-level VQA datasets for renal and gastric cancer to support systematic evaluation. Experiments demonstrate the effectiveness of PlanPath and its compatibility with existing slide-level pathology agents. We hope this work advances pathology agents from single-slide analysis toward more comprehensive patient-level reasoning.

\subsection*{AI use statement}

Generative AI tools were used for language polishing and to assist with dataset construction, including candidate question–answer generation. All AI-assisted content was rigorously reviewed and verified by the authors and domain experts. AI was not used to independently conduct experiments, analyze results, or draw scientific conclusions. The authors take full responsibility for the accuracy, integrity, scientific validity, and final content of this work.



\bibliographystyle{ACM-Reference-Format}
\bibliography{main}
\appendix
\clearpage
\let\PlanPathOriginalIclrRuler\iclrruler
\renewcommand{\iclrruler}[1]{}
\thispagestyle{empty}

\begin{center}
  \vspace*{0.5cm}
  {\LARGE\bfseries PlanPath: Exploration-Guided Orchestration\\
  for Patient-Level Pathology Reasoning\par}
  \vspace{0.45cm}
  {\Large\itshape Supplementary Material\par}
  \vspace{1.0cm}
\end{center}

\begin{center}
  {\large\bfseries Table of Contents\par}
\end{center}
\vspace{0.55em}

\begingroup
  \hypersetup{pdfborder={0 0 0}} 
  \setcounter{tocdepth}{2}
  \renewcommand{\baselinestretch}{1.18}\selectfont
  \setlength{\parskip}{0.18em}
  \makeatletter\@starttoc{apc}\makeatother
\endgroup

\clearpage
\let\iclrruler\PlanPathOriginalIclrRuler

\makeatletter
\let\PlanPathOriginalAddcontentsline\addcontentsline
\renewcommand{\addcontentsline}[3]{%
  \def\PlanPathContentsTarget{#1}%
  \def\PlanPathContentsToc{toc}%
  \ifx\PlanPathContentsTarget\PlanPathContentsToc
    \PlanPathOriginalAddcontentsline{apc}{#2}{#3}%
  \else
    \PlanPathOriginalAddcontentsline{#1}{#2}{#3}%
  \fi
}
\makeatother

\section{Additional Dataset Details}
\label{app:dataset}

\subsection{Patient-Level Pathology VQA Dataset Construction}
\label{app:dataset_construction}

This section supplements Section~\ref{sec:dataset_construction}
with details of the clinical source records, patient-level
organization, question preparation, and evaluation inputs for
PatientVQA-Gastric and PatientVQA-Renal.

\textbf{Clinical records and patient-level organization.}
PatientVQA-Gastric draws on pathology records for early and
advanced gastric cancer, while PatientVQA-Renal draws on records
from patients undergoing partial or radical nephrectomy. The
source records provide diagnostic descriptions and structured
fields for disease classification and pathological staging.
Pathology identifiers link question--answer annotations to
patient-level WSI collections. WSIs sharing a pathology
identifier are grouped into one collection, and slide
filenames are resolved to their stored full paths. All 213
patients in the final datasets match a source record by
pathology identifier and have a corresponding WSI collection.
Consistent with the inclusion scope in the main text, the
reported statistics include only patients with retained
questions and corresponding WSIs.

\textbf{Question generation and verification.}
GPT-5.6 Sol is used to derive candidate question--answer pairs
from the pathology reports, with reference answers grounded in
explicitly documented diagnostic findings. The questions cover
tumor type, grade, local invasion, margin status, lymph node
involvement, and pathological staging. As described in
Section~\ref{sec:dataset_construction}, two pathologists
cross-review the candidate pairs, excluding those with medical
or logical errors, ambiguous wording, insufficient report
support, or inadequate visual evidence in the available WSIs.
Each retained question is linked to the patient's WSI collection
and has a reference answer for evaluation.

\textbf{Question formats and option order.}
The questions comprise yes/no, single-choice, and open-ended short-answer questions. All 501 single-choice questions have four candidate options. We permute their order while preserving the question wording, option contents, and correct answer. The reference answer is stored as the correct option text, independent of its position. The resulting option order is saved in the final question files and remains fixed throughout evaluation.

\textbf{Evaluation samples.}
Each evaluation sample associates a diagnostic question with
the corresponding patient's WSI collection and, where
applicable, its answer options. Depending on the question,
the relevant visual evidence may occur within one slide or
across multiple slides. Pathology reports are used for
question construction and reference answer verification and
are not provided to models during inference. Reference
answers are reserved for evaluation.

\begin{figure}[t]
\centering
\includegraphics[width=\linewidth]{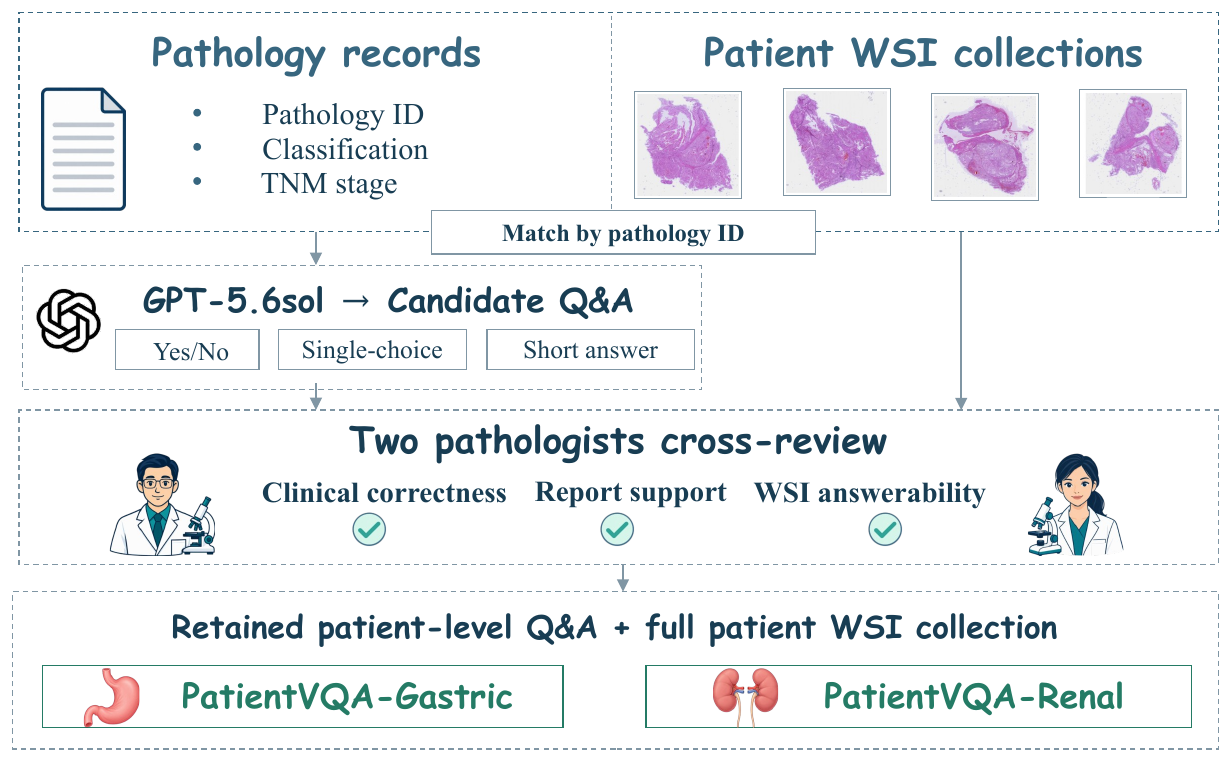}
\caption{Construction of PatientVQA-Gastric and PatientVQA-Renal.
Pathology reports and WSI collections are linked by pathology
identifier. GPT-5.6 Sol generates candidate question--answer
pairs from the reports. Two pathologists cross-review the
candidates for clinical correctness, report support, and
answerability from the available WSIs. Each retained
question--answer pair is associated with the corresponding
patient's full WSI collection.}
\label{fig:app_dataset_pipeline}
\end{figure}

Figure~\ref{fig:app_dataset_pipeline} summarizes the dataset construction process. Table~\ref{tab:app_qa_examples} shows one example of each question type from both datasets. Within each dataset, the
three examples belong to the same patient and are associated with the same WSI collection.

\begin{table}[t]
\centering
\small
\caption{Examples from the final question files. Each dataset's
three examples belong to one patient. Options are shown in their
stored evaluation order. Reference answers are used only for
evaluation.}
\label{tab:app_qa_examples}
\setlength{\tabcolsep}{4pt}
\renewcommand{\arraystretch}{1.12}
\begin{tabular}{p{0.15\textwidth}p{0.63\textwidth}p{0.13\textwidth}}
\toprule
\textbf{Type} & \textbf{Question and options} & \textbf{Answer} \\
\midrule
\multicolumn{3}{l}{\textbf{PatientVQA-Gastric}} \\
Short answer
& What is the deepest level of invasion of the primary tumor in this patient?
& Submucosa \\
Single-choice
& What is the Lauren classification of the primary tumor in this patient?
  \newline \textit{Options:} A: Diffuse type; B: Indeterminate type;
  C: Intestinal type; D: Mixed type.
& Mixed type \\
Yes/No
& Is tumor identified at any submitted margin in this patient?
  \newline \textit{Options:} Yes; No.
& No \\
\midrule
\multicolumn{3}{l}{\textbf{PatientVQA-Renal}} \\
Short answer
& Which pT category is assigned to this patient's primary renal tumor?
& pT1 \\
Single-choice
& What is the highest WHO/ISUP grade of the primary tumor in this patient?
  \newline \textit{Options:} A: Grade 4; B: Grade 2;
  C: Grade 3; D: Grade 1.
& Grade 2 \\
Yes/No
& Does the primary tumor in this patient invade the renal fibrous capsule?
  \newline \textit{Options:} Yes; No.
& No \\
\bottomrule
\end{tabular}
\end{table}

\subsection{Dataset Statistics and Question Distribution}
\label{app:dataset_statistics}

\begin{figure}[t]
\centering
\includegraphics[width=\linewidth]{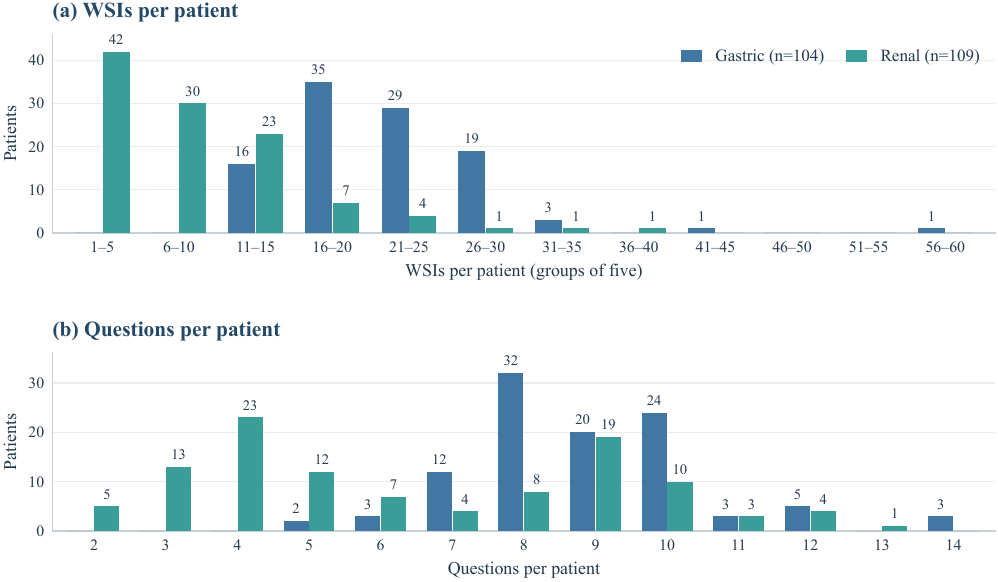}
\caption{Patient-level distributions in PatientVQA-Gastric
and PatientVQA-Renal.
(a) Number of WSIs per patient, grouped into intervals of five.
(b) Number of questions per patient.
Bar heights and the numbers above nonzero bars indicate
patient counts.}
\label{fig:app_dataset_distribution}
\end{figure}

Table~\ref{tab:app_dataset_statistics} summarizes the final
PatientVQA-Gastric and PatientVQA-Renal datasets. Consistent
with Section~\ref{sec:dataset_statistics}, the reported
statistics include only patients with retained questions and
corresponding WSIs. Together, the datasets contain 213
patients, 3,245 WSIs, and 1,623 question--answer pairs.
Figure~\ref{fig:app_dataset_distribution} visualizes the
distributions of WSIs and questions per patient in the two
datasets.

\textbf{Patient-level characteristics.}
The median numbers of WSIs per patient are 21 in Gastric and 8
in Renal, with ranges of 13--58 and 2--36, respectively.
The numbers of questions per patient have medians of 9 and 6,
with ranges of 5--14 and 2--13.
Table~\ref{tab:app_dataset_statistics} also reports the
corresponding means.
Figure~\ref{fig:app_dataset_distribution}(a) shows the
distribution of WSI counts per patient in groups of five,
while Figure~\ref{fig:app_dataset_distribution}(b) shows the
distribution of question counts per patient. Numbers above
the bars indicate the number of patients in each group.

\textbf{Question distribution.}
The datasets contain 595 yes/no, 501 single-choice, and 527
open-ended short-answer questions in total. Their counts and
proportions within each dataset are reported in
Table~\ref{tab:app_dataset_statistics}.
Among single-choice questions, the correct answer appears
at positions A, B, C, and D in 47, 71, 60, and 81 Gastric
questions, respectively, and in 61, 56, 60, and 65 Renal
questions, respectively.

\begin{table}[t]
\centering
\small
\caption{Statistics of PatientVQA-Gastric and PatientVQA-Renal.
Question-type percentages are calculated over all questions
within the corresponding dataset.}
\label{tab:app_dataset_statistics}
\setlength{\tabcolsep}{10pt}
\renewcommand{\arraystretch}{1.10}
\begin{tabular}{lrr}
\toprule
\textbf{Statistic} & \textbf{Gastric} & \textbf{Renal} \\
\midrule
Patients                       & 104           & 109 \\
WSIs                           & 2,227         & 1,018 \\
Question--answer pairs         & 923           & 700 \\
\midrule
WSIs per patient (mean)        & 21.41         & 9.34 \\
WSIs per patient (median)      & 21            & 8 \\
WSIs per patient (range)       & 13--58        & 2--36 \\
Questions per patient (mean)   & 8.88          & 6.42 \\
Questions per patient (median) & 9             & 6 \\
Questions per patient (range)  & 5--14         & 2--13 \\
\midrule
Yes/no                        & 268 (29.0\%)  & 327 (46.7\%) \\
Single-choice                 & 259 (28.1\%)  & 242 (34.6\%) \\
Open-ended short-answer       & 396 (42.9\%)  & 131 (18.7\%) \\
\bottomrule
\end{tabular}
\end{table}

\subsection{Ethics Approval and Data Availability}
\label{app:ethics_data}

All patient-derived data used in this study were collected and processed with approval from the relevant institutional ethics committees and in accordance with applicable ethical and privacy requirements. The question--answer pairs of PatientVQA-Gastric and PatientVQA-Renal are publicly available at \url{https://github.com/JSGe-AI/PlanPath_1}. The original WSIs are available upon request from the corresponding author.

\section{Implementation Details}
\label{app:implementation}

\subsection{Model Parameter Settings}

The model configurations and inference parameters of the main PlanPath modules are summarized in Table~\ref{tab:implementation_details}. The \textbf{Agent Brain} is implemented with Qwen3.5-4B, consisting of 32 layers with a hidden size of 2560, and is responsible for exploration planning, evidence verification, and final answer generation. The \textbf{Morphology Description Tool} employs Patho-R1-7B (Qwen2.5-VL) with a context length of 8,192 tokens and single-image input to generate question-independent morphological descriptions. The \textbf{Slide Router} uses TITAN to extract 768-dimensional WSI embeddings and rank all slides of a patient according to text--WSI semantic similarity. For candidate region retrieval, PLIP encodes each ROI with an input size of $224 \times 224$ and produces 512-dimensional features for semantic relevance ranking.

The \textbf{Tissue-level Analysis Tool} operates on a pre-constructed Tissue Feature Prototype Bank with 768-dimensional feature representations. The softmax temperature for prototype matching is set to 0.04, enabling localization of target tissues, tissue interfaces, and morphologically heterogeneous regions. For cell-level analysis, the \textbf{Cell-level Quantification Tool} employs HistoPLUS with dedicated models at $20\times$ and $40\times$ magnifications. Both models use an input size of $784 \times 784$, with the cell confidence threshold set to 0.50, supporting cell detection, instance segmentation, classification, and density quantification.

To support coarse-to-fine hierarchical evidence acquisition, PlanPath adopts scale-specific WSI cropping and analysis settings at $5\times$, $20\times$, and $40\times$, as summarized in Table~\ref{tab:scale_settings}. At $5\times$, a $4096 \times 4096$ region is cropped from the Level-0 WSI to capture tissue architecture, lesion distribution, and tumor interfaces; HistoPLUS is not activated at this scale. At $20\times$, the Level-0 crop size is reduced to $1024 \times 1024$ for examining cellular organization, epithelial lesions, and local tumor--vessel/nerve relationships. HistoPLUS can be invoked at this scale with a resolution of $0.50~\mu\mathrm{m}/\mathrm{px}$. At $40\times$, a $512 \times 512$ Level-0 region is used for detailed inspection of nuclear and cytoplasmic morphology, tumor subtype, mitotic figures, and subtle suspicious foci, with HistoPLUS operating at $0.25~\mu\mathrm{m}/\mathrm{px}$.

Across all observation scales, the input to Patho-R1 is resized to $512 \times 512$, while the PLIP input is standardized to $224 \times 224$. These scale-specific settings provide a consistent interface across modules while enabling PlanPath to progressively transition from tissue-level structural assessment to local morphological inspection and cell-level quantification during hierarchical evidence acquisition.

\begin{table*}[t]
\centering
\caption{Implementation details of PlanPath.}
\label{tab:implementation_details}

\footnotesize
\renewcommand{\baselinestretch}{1.05}\selectfont
\setlength{\tabcolsep}{4pt}
\renewcommand{\arraystretch}{1.28}

\renewcommand{\tabularxcolumn}[1]{m{#1}}

\begin{tabularx}{\textwidth}{
@{}
>{\raggedright\arraybackslash}m{0.16\textwidth}
>{\raggedright\arraybackslash}m{0.18\textwidth}
>{\raggedright\arraybackslash}m{0.22\textwidth}
>{\raggedright\arraybackslash}X
@{}
}
\toprule

\textbf{Module} &
\textbf{Model / Architecture} &
\textbf{Key Parameters} &
\textbf{Main Function} \\

\midrule

Agent Brain &
Qwen3.5-4B &
32 layers; hidden size 2560 &
Exploration planning, evidence verification, and answer generation. \\

Morphology Description Tool &
Patho-R1-7B (Qwen2.5-VL) &
8,192-token context; single-image input &
Question-independent morphological description. \\

Slide Router &
TITAN &
768-D embeddings &
Text--WSI similarity ranking across patient slides. \\

ROI Retrieval &
PLIP &
CLIP-based; $224 \times 224$ input; 512-D features &
Semantic ranking of candidate ROIs at $5\times$ and above. \\

Tissue-level Analysis Tool &
Tissue Feature Prototype Bank &
768-D features; softmax temperature 0.04 &
Localization of target tissues, interfaces, and heterogeneous regions. \\

Cell-level Quantification Tool &
HistoPLUS &
$20\times$/$40\times$ models; $784 \times 784$ input; confidence thr=0.5 &
Cell detection, instance segmentation, classification, and density quantification. \\

\bottomrule
\end{tabularx}
\end{table*}

\begin{table*}[t]
\centering
\caption{Scale-specific settings for hierarchical evidence acquisition in PlanPath.}
\label{tab:scale_settings}

\footnotesize
\setlength{\tabcolsep}{4pt}
\renewcommand{\arraystretch}{1.25}
\renewcommand{\tabularxcolumn}[1]{m{#1}}

\begin{tabularx}{\textwidth}{
@{}
>{\centering\arraybackslash}m{0.05\textwidth}
>{\centering\arraybackslash}m{0.12\textwidth}
>{\centering\arraybackslash}m{0.115\textwidth}
>{\centering\arraybackslash}m{0.095\textwidth}
>{\raggedright\arraybackslash}X
>{\raggedright\arraybackslash}m{0.145\textwidth}
@{}
}
\toprule

\textbf{Scale} &
\shortstack{\textbf{WSI Level-0}\\\textbf{Crop (px)}} &
\shortstack{\textbf{Patho-R1}\\\textbf{Input (px)}} &
\shortstack{\textbf{PLIP}\\\textbf{Input (px)}} &
\textbf{Main Purpose} &
\textbf{HistoPLUS} \\

\midrule

$5\times$ &
$4096 \times 4096$ &
$512 \times 512$ &
$224 \times 224$ &
Tissue architecture, lesion distribution, and tumor interfaces. &
Not used. \\

$20\times$ &
$1024 \times 1024$ &
$512 \times 512$ &
$224 \times 224$ &
Cellular organization, epithelial lesions, and tumor--vessel/nerve relations. &
$0.50~\mu$m/px; $784 \times 784$ input. \\

$40\times$ &
$512 \times 512$ &
$512 \times 512$ &
$224 \times 224$ &
Nuclear/cytoplasmic morphology, tumor subtype, mitoses, and subtle foci. &
$0.25~\mu$m/px; $784 \times 784$ input. \\

\bottomrule
\end{tabularx}
\end{table*}


\subsection{Implementation of Slide Router}
\label{app:slide_router}

\paragraph{Input, Output, and Retrieval Description Construction.}
The Slide Router is used only to determine the priority order in which the WSIs of the current patient are examined. Its outputs consist of the ranked slides and their corresponding routing scores, which are not treated as histopathological evidence. The inputs include all WSIs from the same patient and the slide-level retrieval descriptions generated by the Agent Brain for the current question.

For each question, the Agent Brain generates $1$--$3$ concise English caption-style descriptions that characterize the tissue appearances that should be preferentially searched for at the slide level. These descriptions may include dataset-level organ context, but do not contain slide filenames, specimen-site labels, or reference answers. For Yes/No questions, the retrieval descriptions cover both the target finding and an assessable contrasting state. For MCQ questions, they cover the major morphological axes associated with different candidate answers, thereby avoiding a retrieval direction dominated by a single option. All retrieval descriptions are used only for slide ranking and do not directly participate in downstream answer generation.

\paragraph{Slide and Text Representations.}
Each WSI is encoded by the Slide Router into a $768$-dimensional global slide representation,
$\mathbf{v}_i \in \mathbb{R}^{768}$.
In practice, the visual representation of each slide is computed only once and stored in single precision, avoiding repeated WSI encoding for different questions from the same patient. After loading, each representation is checked for dimensional validity, numerical finiteness, and non-zero norm, and is then re-normalized using $L_2$ normalization.

The retrieval descriptions are processed by a text encoder aligned with the visual encoder in the same semantic embedding space. The text encoder uses a context window of length $128$, with the final position reserved for the global representation token. Inference is performed on GPU in half precision, and the resulting text representations are converted to single precision and re-normalized with $L_2$ normalization. Let $\mathbf{t}_j$ denote the representation of the $j$-th retrieval description. The normalized slide and text representations are defined as
\[
\hat{\mathbf{v}}_i
=
\frac{\mathbf{v}_i}{\lVert \mathbf{v}_i \rVert_2},
\qquad
\hat{\mathbf{t}}_j
=
\frac{\mathbf{t}_j}{\lVert \mathbf{t}_j \rVert_2}.
\]

\paragraph{Similarity Computation and Patient-level Ranking.}
The Slide Router performs retrieval and ranking only within the slide collection of the current patient. For slide $S_i$ and retrieval description $d_j$, the matching score is defined as the cosine similarity between their normalized representations:
\[
c_{ij}
=
\hat{\mathbf{v}}_i^{\top}\hat{\mathbf{t}}_j.
\]

When multiple retrieval descriptions are generated for a question, the final routing score of slide $S_i$ is defined as
\[
r_i
=
\max_{1 \leq j \leq M} c_{ij},
\qquad
M \in \{1,2,3\}.
\]
The max aggregation treats the retrieval descriptions as complementary morphological hypotheses, ensuring that a slide strongly matching any one target pattern is not suppressed by low similarity to the remaining descriptions. In addition to the final routing score, the system records the retrieval description producing the maximum score and the similarity scores against all descriptions for auditability.

All slides of the current patient are ranked in descending order according to $r_i$. When two slides have identical routing scores, the anonymized slide ID is used as a deterministic tie-breaker to ensure reproducible ranking across repeated runs.

\paragraph{Candidate Slide Pool Construction.}
By default, the top $5$ analyzable slides form the initial candidate pool. The pool size is further adjusted according to the evidence requirements of the current question.

For extremum-seeking tasks or focal spatial-interface tasks, the candidate pool is expanded to include at least the top $8$ slides. For coverage-sensitive tasks that require sufficient support for patient-level negative, extremum, or counting conclusions, all analyzable slides are retained. Examples include assessment of the deepest gastric wall invasion and pT stage, perineural or lymphovascular invasion, renal capsular involvement, and regional lymph-node status or pN stage.

Candidate slides are subsequently processed by ROI-level analysis in routing order. For presence-detection questions, exploration may terminate early once direct positive evidence passes the Evidence Gate. In contrast, tasks requiring sufficient candidate-pool coverage continue to inspect the remaining slides until the predefined coverage and stopping criteria are satisfied.

A slide is included in the effective candidate pool only when its global representation, original WSI, and the data required for downstream ROI analysis are all accessible. Slides missing a global representation cannot be assigned a routing score. Slides that can be ranked at the slide level but lack the original image or ROI-analysis data are marked as unanalyzable and excluded from the candidate pool. All missing or exclusion states are recorded in the execution log. If all slides of a patient lack valid global representations, the Slide Router terminates the routing process and returns an error state.

\paragraph{Caching and Reproducibility.}
To reduce repeated computation across multiple questions from the same patient, the system caches text representations, slide representations, and complete ranking results. The text-cache key is determined by the normalized set of retrieval descriptions. The slide-representation cache is bound to the corresponding representation file, while the persistent ranking cache is further conditioned on the current question, Evidence Plan, model state, and slide state. Any change in the relevant inputs triggers recomputation.

The caching mechanism only reuses deterministic intermediate results under identical inputs. It introduces no additional clinical information and does not modify the candidate-pool construction or stopping rules. Regardless of whether a ranking is obtained through real-time computation or cache reuse, the similarity scores, ranking positions, and matched retrieval descriptions produced by the Slide Router are used solely to determine the subsequent slide exploration order. They are not treated as diagnostic evidence and are not directly propagated to final answer generation.

\subsection{Implementation of Tissue-Level Analysis Tool}
\label{app:tissue_analysis}


\paragraph{Module Design.}
The Tissue-level Analysis Tool is a weakly supervised spatial analysis module based on tissue feature prototypes. Rather than directly diagnosing an ROI, it converts a WSI into a spatial tissue map with soft class probabilities, providing a unified representation for target-region retrieval, tissue composition quantification, and tissue-interface localization. The module consists of three stages: tissue prototype bank construction, patch-level prototype matching, and region-level spatial aggregation.

The prototype bank is constructed once during development. During inference, patch representations are extracted from the current WSI and matched against the fixed prototype bank to generate a soft tissue map. Question-independent patch representations and tissue maps are cached to avoid redundant computation across multiple questions from the same slide.

\paragraph{Tissue Prototype Bank Construction.}
The gastric prototype bank was constructed from \textbf{TCGA-STAD}, comprising 1,051 WSIs and 1,944,093 pathology patch representations of 768 dimensions. The renal prototype bank was constructed from \textbf{TCGA-RCC}, comprising 934 WSIs and 3,408,240 patch representations. All patch representations were first $L_2$-normalized and then clustered using MiniBatch $K$-means with $K=100$. The mini-batch size was set to 8,192, with three initializations and a fixed random seed of 42.

For each cluster, the 30 high-resolution patches closest to the cluster center were manually reviewed to determine the dominant tissue category, morphological purity, and secondary tissue components. After review, 11 tissue categories were retained for TCGA-STAD and 8 for TCGA-RCC. Because hemorrhagic regions in TCGA-STAD did not form an independent dominant cluster, 16 additional patches with a purity of at least 0.90 were selected from mixed clusters to supplement the hemorrhage category.

\paragraph{Representative Prototype Selection.}
For each reviewed cluster, the 10 patches closest to the cluster center were selected as representative prototypes, while limiting each WSI to at most two selected patches per cluster to reduce slide-specific bias. Randomly sampled and boundary patches were used only for cluster review and were not included in the final prototype bank.

The resulting banks contained 1,016 representative patches for TCGA-STAD and 1,000 for TCGA-RCC. Before inference, representations belonging to the same cluster--tissue-category pair were averaged and re-normalized, yielding 102 inference prototypes for TCGA-STAD and 100 for TCGA-RCC. This design reduces noise from individual patches while preserving intra-class morphological diversity across different clusters. Detailed prototype statistics are summarized in Table~\ref{tab:tissue_prototype_statistics}.

\begin{table*}[t]
\centering
\caption{Statistics of the tissue feature prototype banks used by the Tissue-level Analysis Tool.}
\label{tab:tissue_prototype_statistics}

\scriptsize
\setlength{\tabcolsep}{3.2pt}
\renewcommand{\arraystretch}{1.08}
\renewcommand{\tabularxcolumn}[1]{m{#1}}

\begin{tabularx}{\textwidth}{
@{}
>{\centering\arraybackslash}m{0.085\textwidth}
>{\centering\arraybackslash}m{0.055\textwidth}
>{\centering\arraybackslash}m{0.075\textwidth}
>{\centering\arraybackslash}m{0.035\textwidth}
>{\raggedright\arraybackslash}X
>{\centering\arraybackslash}m{0.085\textwidth}
>{\centering\arraybackslash}m{0.105\textwidth}
>{\centering\arraybackslash}m{0.105\textwidth}
@{}
}
\toprule

\textbf{Cohort} &
\textbf{\#WSIs} &
\textbf{\#Patches} &
\textbf{$K$} &
\textbf{Tissue Category} &
\shortstack{\textbf{Dominant}\\\textbf{Clusters}} &
\shortstack{\textbf{Prototype}\\\textbf{Patches}} &
\shortstack{\textbf{Inference}\\\textbf{Prototypes}} \\

\midrule

\multirow[c]{12}{*}{TCGA-STAD}
& \multirow[c]{12}{*}{1,051}
& \multirow[c]{12}{*}{1,944,093}
& \multirow[c]{12}{*}{100}
& Artifact / background
& 15 & 150 & 15 \\

&
&
&
& Non-tumor mucosa / epithelium
& 3 & 30 & 3 \\

&
&
&
& Gland-forming tumor
& 21 & 210 & 21 \\

&
&
&
& Diffuse / poorly cohesive tumor
& 16 & 160 & 16 \\

&
&
&
& Desmoplastic stroma / tumor bed
& 19 & 190 & 19 \\

&
&
&
& Immune / lymphoid tissue
& 3 & 30 & 3 \\

&
&
&
& Necrosis / ulcer debris
& 1 & 10 & 1 \\

&
&
&
& Mucin pool
& 1 & 10 & 1 \\

&
&
&
& Hemorrhage / blood$^{\dagger}$
& 0 & 16 & 2 \\

&
&
&
& Adipose / perigastric soft tissue
& 8 & 80 & 8 \\

&
&
&
& Muscularis
& 13 & 130 & 13 \\

\cmidrule(lr){5-8}

&
&
&
& \textit{Total}
& 100 & 1,016 & 102 \\

\midrule

\multirow[c]{9}{*}{TCGA-RCC}
& \multirow[c]{9}{*}{934}
& \multirow[c]{9}{*}{3,408,240}
& \multirow[c]{9}{*}{100}
& Artifact / background
& 18 & 180 & 18 \\

&
&
&
& Viable renal tumor
& 36 & 360 & 36 \\

&
&
&
& Normal renal parenchyma
& 6 & 60 & 6 \\

&
&
&
& Interstitium / fibrosis
& 23 & 230 & 23 \\

&
&
&
& Immune / lymphoid tissue
& 2 & 20 & 2 \\

&
&
&
& Necrosis / debris
& 3 & 30 & 3 \\

&
&
&
& Hemorrhage / blood
& 8 & 80 & 8 \\

&
&
&
& Adipose / soft tissue
& 4 & 40 & 4 \\

\cmidrule(lr){5-8}

&
&
&
& \textit{Total}
& 100 & 1,000 & 100 \\

\bottomrule
\end{tabularx}

\vspace{2pt}

\parbox{0.98\textwidth}{
\scriptsize
\textit{Note:}
Dominant Clusters denotes the number of manually reviewed clusters assigned primarily to each tissue category.
Prototype Patches denotes the representative patches retained in the prototype bank, while Inference Prototypes denotes the prototypes obtained by averaging representations within each cluster--category group.
$^{\dagger}$The TCGA-STAD hemorrhage category was supplemented with 16 high-purity patches from two mixed source clusters; therefore, it has no independent dominant cluster but yields two inference prototypes.
}

\end{table*}

\paragraph{Patch-level Tissue Matching.}
For a patch representation $\mathbf{z}_i$ from the target WSI and the $k$-th normalized prototype $\mathbf{p}_{c,k}$ of tissue category $c$, the category score is defined as the maximum cosine similarity over all prototypes belonging to that category:
\[
s_{i,c}
=
\max_k
\hat{\mathbf{z}}_i^{\top}
\hat{\mathbf{p}}_{c,k}.
\]
This within-class maximum allows different morphological subtypes of the same tissue category to be represented by separate prototypes.

The category scores are converted into soft probabilities using a temperature parameter $\tau=0.04$:
\[
q_{i,c}
=
\frac{
\exp\left[(s_{i,c}-\max_{c'}s_{i,c'})/\tau\right]
}{
\sum_{c'}
\exp\left[(s_{i,c'}-\max_{c''}s_{i,c''})/\tau\right]
}.
\]

A patch is marked as \textit{unknown} when the highest class probability is below 0.30 or when the raw similarity margin between the highest- and second-highest-scoring classes is below 0.01. For each patch, the system records its spatial coordinates, soft class probabilities, predicted class, confidence, class margin, normalized entropy, and unknown status, thereby constructing a soft tissue map in the original WSI coordinate system.

\paragraph{Regional Quantification and Interface Analysis.}
For a candidate region $R$, the Tissue-level Analysis Tool aggregates the soft probabilities of all intersecting patches to estimate tissue composition, target-tissue scores, mean confidence, unknown ratio, and tissue heterogeneity.

For whole-slide quantification, the soft area of tissue category $c$ is defined as
\[
A_c
=
\sum_{i\in\mathcal{I}_{\mathrm{known}}}
a_i q_{i,c},
\]
and its relative tissue fraction is
\[
F_c
=
\frac{A_c}
{\sum_{c'} A_{c'}},
\]
where $a_i$ denotes the area of patch $i$ in the Level-0 coordinate system and $\mathcal{I}_{\mathrm{known}}$ denotes patches not labeled as unknown. The proportion of known tissue is additionally used as a quality-control criterion before reporting tissue fractions.

For a tissue pair $(c_1,c_2)$, interface candidates are identified by combining their soft-probability co-occurrence with spatial proximity. High-scoring neighboring patches from different tissue categories are localized, and the midpoint of the best-matching pair is used as a candidate interface center for subsequent ROI relocation and higher-magnification inspection. For slide-edge analysis, the tissue map is rasterized and boundary-connected background regions are identified using flood fill, allowing true peripheral regions to be distinguished from internal clefts or artifacts.

The resulting tissue composition, proportion, and interface information is used for ROI ranking, spatial relocation, and explicit tissue-level quantification. Importantly, the predicted tissue classes and probabilities are not treated as direct evidence for tumor subtype, grade, stage, or invasion; such conclusions require confirmation through subsequent morphological observation.

\subsection{Baseline Settings}

For a fair comparison, all baseline methods follow their official implementations and recommended inference settings unless otherwise specified. We use Qwen3.5-4B as the unified model for patient-level output aggregation across all applicable baselines. For slide-level pathology agents, Qwen3.5-4B is additionally used for within-slide evidence integration and final slide-level answer generation, ensuring a consistent reasoning backbone across agentic baselines. Each WSI is first processed independently according to the corresponding baseline pipeline, after which the resulting slide-level outputs are aggregated by Qwen3.5-4B into a single patient-level answer. Apart from these unified integration and aggregation settings, all model architectures, visual encoders, preprocessing procedures, tool configurations, and inference parameters follow the official baseline implementations.


\section{Additional Experimental Results}
\label{app:additional_results}

\subsection{Detailed Results on Patient-Level Benchmarks}
\label{app:detailed_results}

The main text reports a compact comparison on PatientVQA-Gastric and PatientVQA-Renal using Total, MCQ, Yes/No, and Open accuracy.
Tables~\ref{tab:gastric_detailed} and~\ref{tab:renal_detailed} provide the complete per-dataset results and additionally report BLEU-1 and ROUGE-L for open-ended questions.

On PatientVQA-Gastric, PlanPath achieves a BLEU-1 score of 46.66 and a ROUGE-L score of 51.79, exceeding the strongest baselines by 17.98 and 7.47 percentage points, respectively. On PatientVQA-Renal, the corresponding scores reach 54.87 and 59.75, with improvements of 23.48 and 30.38 points. Together with the accuracy results reported in the main text, these results
show that the improvement of PlanPath on open-ended questions is consistent across both exact-match and text-overlap metrics.

\begin{table*}[t]
\centering
\caption{
Detailed performance on PatientVQA-Gastric.
BLEU-1 and ROUGE-L are additionally reported for open-ended questions.
All values are reported as percentages.
The best and second-best results in each column are shown in
\textbf{bold} and \underline{underlined}, respectively.
}
\label{tab:gastric_detailed}

\setlength{\tabcolsep}{6pt}
\renewcommand{\arraystretch}{1.08}

\resizebox{\textwidth}{!}{
\begin{tabular}{lccccccc}
\toprule
\textbf{Method}
& \textbf{Input}
& \textbf{Total}
& \textbf{MCQ}
& \textbf{Yes/No}
& \textbf{Open}
& \textbf{BLEU-1}
& \textbf{ROUGE-L} \\
\midrule

\multicolumn{8}{l}{\textbf{\textit{Pretrained MLLM Models}}} \\

GPT-5.5 \citep{openai2026gpt55}
& Slide-T
& 35.10
& 39.77
& 30.97
& 34.85
& 28.22
& 36.84 \\

GPT-5.6-Luna \citep{openai2026gpt56}
& Slide-T
& 37.70
& 36.29
& 50.75
& 29.80
& 18.65
& 27.70 \\

Qwen3.5-4B \citep{qwen2026qwen35}
& Slide-T
& 50.38
& 42.86
& 65.30
& \underline{45.20}
& \underline{28.68}
& 44.11 \\

Qwen3.5-9B \citep{qwen2026qwen35}
& Slide-T
& 47.24
& 41.70
& 62.31
& 40.66
& 25.95
& \underline{44.32} \\

\midrule
\multicolumn{8}{l}{\textbf{\textit{Supervised Pathology Models}}} \\

MedDr \citep{MedDr}
& Slide-T
& 30.44
& 30.50
& 64.18
& 7.58
& 2.85
& 5.77 \\

WSI-VQA \citep{wsi-vqa}
& WSI
& 32.83
& 44.79
& 47.39
& 15.15
& 14.02
& 12.39 \\

TITAN \citep{TITAN}
& WSI
& --
& 26.64
& 39.93
& --
& --
& -- \\

Quilt-LLaVA \citep{quilt-LLAVA}
& Patch
& 29.47
& 51.35
& 35.82
& 10.86
& 4.64
& 13.27 \\

WSI-LLaVA \citep{Wsi-llava}
& WSI
& 32.50
& 22.78
& \underline{67.91}
& 14.90
& 20.93
& 20.49 \\

MedGemma-1.5-4B \citep{medgemma}
& Slide-T
& \underline{50.81}
& \underline{52.12}
& 64.93
& 40.40
& 10.44
& 15.01 \\

SlideChat \citep{slidechat}
& WSI
& 34.78
& 26.25
& 63.43
& 20.96
& 10.32
& 15.82 \\

\midrule
\multicolumn{8}{l}{\textbf{\textit{Agentic Systems}}} \\

PathAgent \citep{chen2025pathagent}
& WSI
& 33.91
& 40.93
& 38.81
& 26.01
& 13.80
& 14.44 \\

PathNavigate \citep{pathnavigate}
& WSI
& 34.45
& 43.24
& 58.96
& 12.12
& 8.25
& 9.07 \\

GIANT \citep{giant}
& WSI
& 37.27
& 44.40
& 33.21
& 35.35
& 14.55
& 31.33 \\

BEACON \citep{Beacon}
& WSI
& --
& 33.98
& 38.81
& --
& --
& -- \\

\rowcolor{gray!10}
\textbf{PlanPath (ours)}
& WSI
& \textbf{60.24}
& \textbf{57.92}
& \textbf{70.15}
& \textbf{55.05}
& \textbf{46.66}
& \textbf{51.79} \\

\bottomrule
\end{tabular}
}

\end{table*}

\begin{table*}[t]
\centering
\caption{
Detailed performance on PatientVQA-Renal.
BLEU-1 and ROUGE-L are additionally reported for open-ended questions.
All values are reported as percentages.
The best and second-best results in each column are shown in
\textbf{bold} and \underline{underlined}, respectively.
}
\label{tab:renal_detailed}

\setlength{\tabcolsep}{6pt}
\renewcommand{\arraystretch}{1.08}

\resizebox{\textwidth}{!}{
\begin{tabular}{lccccccc}
\toprule
\textbf{Method}
& \textbf{Input}
& \textbf{Total}
& \textbf{MCQ}
& \textbf{Yes/No}
& \textbf{Open}
& \textbf{BLEU-1}
& \textbf{ROUGE-L} \\
\midrule

\multicolumn{8}{l}{\textbf{\textit{Pretrained MLLM Models}}} \\

GPT-5.5 \citep{openai2026gpt55}
& Slide-T
& 40.71
& 23.97
& 60.55
& 22.14
& 17.86
& 19.86 \\

GPT-5.6-Luna \citep{openai2026gpt56}
& Slide-T
& 49.29
& 47.11
& 62.08
& 21.37
& 24.75
& 23.85 \\

Qwen3.5-4B \citep{qwen2026qwen35}
& Slide-T
& 55.14
& 50.00
& 69.11
& 29.77
& 21.67
& 20.94 \\

Qwen3.5-9B \citep{qwen2026qwen35}
& Slide-T
& 55.57
& 46.69
& 71.56
& \underline{32.06}
& \underline{31.39}
& 27.19 \\

\midrule
\multicolumn{8}{l}{\textbf{\textit{Supervised Pathology Models}}} \\

MedDr \citep{MedDr}
& Slide-T
& 55.86
& 58.26
& 70.34
& 15.27
& 16.76
& 9.78 \\

WSI-VQA \citep{wsi-vqa}
& WSI
& 15.86
& 13.22
& 20.18
& 9.92
& 1.72
& 0.93 \\

TITAN \citep{TITAN}
& WSI
& --
& 13.64
& 57.80
& --
& --
& -- \\

Quilt-LLaVA \citep{quilt-LLAVA}
& Patch
& 37.00
& \underline{61.98}
& 29.97
& 8.40
& 9.87
& 10.15 \\

WSI-LLaVA \citep{Wsi-llava}
& WSI
& 49.00
& 35.12
& 71.87
& 17.56
& 21.24
& 16.09 \\

MedGemma-1.5-4B \citep{medgemma}
& Slide-T
& \underline{60.57}
& 59.50
& \underline{73.39}
& 30.53
& 25.26
& \underline{29.37} \\

SlideChat \citep{slidechat}
& WSI
& 44.43
& 17.77
& 71.56
& 25.95
& 9.41
& 14.32 \\

\midrule
\multicolumn{8}{l}{\textbf{\textit{Agentic Systems}}} \\

PathAgent \citep{chen2025pathagent}
& WSI
& 33.14
& 46.28
& 30.28
& 16.03
& 23.89
& 19.84 \\

PathNavigate \citep{pathnavigate}
& WSI
& 55.14
& 57.85
& 67.58
& 19.08
& 21.46
& 25.38 \\

GIANT \citep{giant}
& WSI
& 47.14
& 34.30
& 63.30
& 30.53
& 16.50
& 21.11 \\

BEACON \citep{Beacon}
& WSI
& --
& 47.11
& 25.08
& --
& --
& -- \\

\rowcolor{gray!10}
\textbf{PlanPath (ours)}
& WSI
& \textbf{72.71}
& \textbf{78.51}
& \textbf{74.92}
& \textbf{56.49}
& \textbf{54.87}
& \textbf{59.75} \\

\bottomrule
\end{tabular}
}

\end{table*}



\begin{table*}[t]
\centering
\caption{
Performance on PatientVQA-Gastric using majority-voting aggregation for
slide-level baselines. PlanPath is shown unchanged for reference; results
using Qwen3.5-4B aggregation are reported in the main-text
Table~\ref{tab:main_results}. Best and second-best results are shown in
\textbf{bold} and \underline{underlined}, respectively.
}
\label{tab:aggregation_gastric}

\setlength{\tabcolsep}{6pt}
\renewcommand{\arraystretch}{1.08}

\resizebox{\textwidth}{!}{
\begin{tabular}{lccccccc}
\toprule
\textbf{Method}
& \textbf{Input}
& \textbf{Total}
& \textbf{MCQ}
& \textbf{Yes/No}
& \textbf{Open}
& \textbf{BLEU-1}
& \textbf{ROUGE-L} \\
\midrule

\multicolumn{8}{l}{\textbf{\textit{Pretrained MLLM Models}}} \\

GPT-5.5 \citep{openai2026gpt55}
& Slide-T & 45.50 & 45.17 & 46.27 & 45.20 & 24.32 & 22.05 \\

GPT-5.6-Luna \citep{openai2026gpt56}
& Slide-T & 43.45 & 41.31 & 64.55 & 30.56 & 21.19 & 26.45 \\

Qwen3.5-4B \citep{qwen2026qwen35}
& Slide-T & \underline{53.30} & 47.49 & 64.18
& \underline{49.75} & 34.64 & 50.09 \\

Qwen3.5-9B \citep{qwen2026qwen35}
& Slide-T & 51.25 & 46.33 & 64.18 & 45.71
& \underline{35.43} & \underline{51.52} \\

\midrule
\multicolumn{8}{l}{\textbf{\textit{Supervised Pathology Models}}} \\

MedDr \citep{MedDr}
& Slide-T & 34.99 & 35.91 & 64.18 & 14.65 & 5.97 & 12.79 \\

WSI-VQA \citep{wsi-vqa}
& WSI & 30.01 & 37.45 & 49.63 & 11.87 & 13.93 & 12.32 \\

TITAN \citep{TITAN}
& WSI & -- & 33.20 & 45.90 & -- & -- & -- \\

Quilt-LLaVA \citep{quilt-LLAVA}
& Patch & 29.47 & 51.35 & 35.82 & 10.86 & 4.96 & 13.04 \\

WSI-LLaVA \citep{Wsi-llava}
& WSI & 32.18 & 22.78 & 69.03 & 13.38 & 23.08 & 20.49 \\

MedGemma-1.5-4B \citep{medgemma}
& Slide-T & 44.85 & \underline{51.74} & 64.18
& 27.27 & 12.46 & 15.01 \\

SlideChat \citep{slidechat}
& WSI & 35.32 & 27.80 & 64.18 & 20.71 & 8.37 & 15.45 \\

\midrule
\multicolumn{8}{l}{\textbf{\textit{Agentic Systems}}} \\

PathAgent \citep{chen2025pathagent}
& WSI & 40.85 & 41.70 & 51.87 & 32.83 & 17.70 & 16.02 \\

PathNavigate \citep{pathnavigate}
& WSI & 42.04 & 47.88 & \textbf{79.48} & 12.88 & 2.96 & 3.40 \\

GIANT \citep{giant}
& WSI & 47.45 & 47.49 & 60.07 & 38.89 & 25.52 & 38.81 \\

BEACON \citep{Beacon}
& WSI & -- & 36.68 & 46.27 & -- & -- & -- \\

\rowcolor{gray!10}
\textbf{PlanPath (ours)}
& WSI
& \textbf{60.24}
& \textbf{57.92}
& \underline{70.15}
& \textbf{55.05}
& \textbf{46.66}
& \textbf{51.79} \\

\bottomrule
\end{tabular}
}

\vspace{-3mm}
\end{table*}
\begin{table*}[t]
\centering
\caption{
Performance on PatientVQA-Renal using majority-voting aggregation for
slide-level baselines. PlanPath is shown unchanged for reference; results
using Qwen3.5-4B aggregation are reported in the main-text
Table~\ref{tab:main_results}. Best and second-best results are shown in
\textbf{bold} and \underline{underlined}, respectively.
}
\label{tab:aggregation_renal}

\setlength{\tabcolsep}{6pt}
\renewcommand{\arraystretch}{1.08}

\resizebox{\textwidth}{!}{
\begin{tabular}{lccccccc}
\toprule
\textbf{Method}
& \textbf{Input}
& \textbf{Total}
& \textbf{MCQ}
& \textbf{Yes/No}
& \textbf{Open}
& \textbf{BLEU-1}
& \textbf{ROUGE-L} \\
\midrule

\multicolumn{8}{l}{\textbf{\textit{Pretrained MLLM Models}}} \\

GPT-5.5 \citep{openai2026gpt55}
& Slide-T & 46.29 & 35.95 & 62.69 & 24.43 & 22.12 & 21.69 \\

GPT-5.6-Luna \citep{openai2026gpt56}
& Slide-T & 55.00 & \underline{62.81} & 63.30
& 19.85 & 24.54 & 21.62 \\

Qwen3.5-4B \citep{qwen2026qwen35}
& Slide-T & 55.57 & 48.76 & 70.64 & 30.53 & 24.77 & 21.87 \\

Qwen3.5-9B \citep{qwen2026qwen35}
& Slide-T & 54.29 & 41.74 & 71.87
& \underline{33.59} & \underline{32.71} & \underline{28.72} \\

\midrule
\multicolumn{8}{l}{\textbf{\textit{Supervised Pathology Models}}} \\

MedDr \citep{MedDr}
& Slide-T & 56.00 & 59.50 & 70.03 & 14.50 & 15.37 & 8.98 \\

WSI-VQA \citep{wsi-vqa}
& WSI & 18.14 & 13.64 & 25.08 & 9.16 & 0.87 & 0.56 \\

TITAN \citep{TITAN}
& WSI & -- & 16.12 & 63.61 & -- & -- & -- \\

Quilt-LLaVA \citep{quilt-LLAVA}
& Patch & 36.71 & 61.98 & 29.97 & 6.87 & 9.62 & 8.66 \\

WSI-LLaVA \citep{Wsi-llava}
& WSI & 55.86 & 50.41 & \underline{75.23}
& 17.56 & 20.35 & 13.80 \\

MedGemma-1.5-4B \citep{medgemma}
& Slide-T & 60.00 & 61.16 & 71.56 & 29.01 & 22.67 & 27.02 \\

SlideChat \citep{slidechat}
& WSI & 44.00 & 18.18 & 70.64 & 25.19 & 9.68 & 13.04 \\

\midrule
\multicolumn{8}{l}{\textbf{\textit{Agentic Systems}}} \\

PathAgent \citep{chen2025pathagent}
& WSI & 37.86 & 51.65 & 36.09 & 16.79 & 19.42 & 16.63 \\

PathNavigate \citep{pathnavigate}
& WSI & \underline{60.43} & 59.09 & \textbf{77.06}
& 21.37 & 23.79 & 26.47 \\

GIANT \citep{giant}
& WSI & 51.14 & 40.08 & 67.58 & 30.53 & 17.41 & 19.94 \\

BEACON \citep{Beacon}
& WSI & -- & 51.65 & 22.63 & -- & -- & -- \\

\rowcolor{gray!10}
\textbf{PlanPath (ours)}
& WSI
& \textbf{72.71}
& \textbf{78.51}
& 74.92
& \textbf{56.49}
& \textbf{54.87}
& \textbf{59.75} \\

\bottomrule
\end{tabular}
}

\vspace{-3mm}
\end{table*}

\subsection{Comparison of Baseline Aggregation Strategies}
\label{app:aggregation}

To examine whether baseline performance depends on how independently generated slide-level predictions are combined, we compare majority voting with Qwen3.5-4B aggregation, which is used for the baseline results in the main text. Both strategies operate on the same saved slide-level predictions, while the underlying visual inference remains unchanged.

\paragraph{Majority Voting.}
Each non-empty and parseable slide-level answer contributes one vote, and the most frequent answer is selected as the patient-level prediction. For open-ended questions, model-specific parsers extract concise answers by removing answer prefixes and truncating at predefined sentence or punctuation boundaries. The extracted answers are normalized for case, whitespace, hyphens, and predefined grade expressions before identical answers are grouped. Empty or unparseable outputs are excluded. If no valid answer remains, the patient-level prediction is set to \texttt{None} and counted as incorrect. Ties are resolved using prediction confidence when available, followed by a deterministic model-specific rule if necessary.

\paragraph{Qwen3.5-4B Aggregation.}
Qwen3.5-4B receives the question, answer options when applicable, and all valid slide-level answers from the same patient, and selects a single patient-level answer. For open-ended questions, it is instructed to return the selected answer verbatim without paraphrasing, merging, or expanding it. The aggregation model receives no images, pathology reports, or reference answers.

\paragraph{Results.}
Tables~\ref{tab:aggregation_gastric} and~\ref{tab:aggregation_renal} report the results using majority voting, while the corresponding Qwen3.5-4B aggregation results are reported in the main-text Table~\ref{tab:main_results}. The effect of aggregation varies across baselines, with majority voting improving some methods while reducing others. On PatientVQA-Gastric, the strongest baseline under majority voting achieves 53.30\% Total Accuracy, compared with 60.24\% for PlanPath, corresponding to a 6.94 percentage-point difference. Under Qwen3.5-4B aggregation, the strongest baseline achieves 50.81\%, corresponding to a 9.43-point difference. On PatientVQA-Renal, the strongest baseline reaches 60.43\% with majority voting and 60.57\% with Qwen3.5-4B aggregation, compared with 72.71\% for PlanPath, yielding differences of 12.28 and 12.14 percentage points, respectively. These results show that the performance advantage of PlanPath is maintained under both baseline aggregation strategies.

\begin{figure*}[t]
    \centering
    \includegraphics[width=\textwidth]{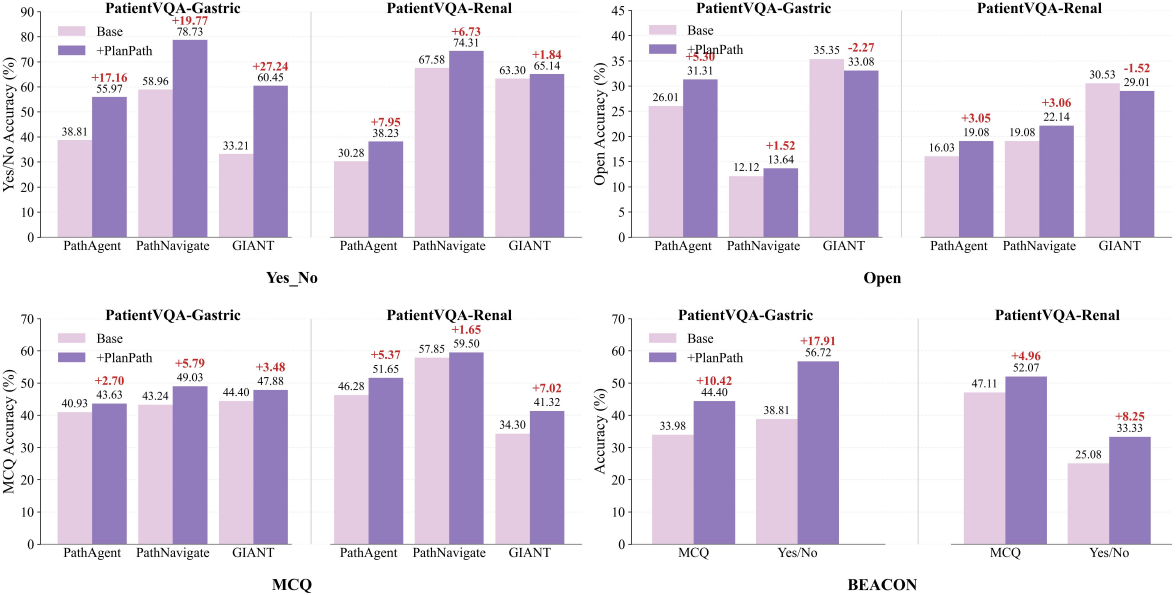}
    \caption{
    Detailed compatibility results of PlanPath with different slide-level pathology agents on PatientVQA-Gastric and PatientVQA-Renal.
    For PathAgent, PathNavigate, and GIANT, results are reported separately for MCQ, Yes/No, and Open questions.
    BEACON is evaluated on MCQ and Yes/No questions.
    Each pair of bars compares the original agent (Base) with the same agent integrated into PlanPath (+PlanPath), and the red annotations indicate the absolute performance change in percentage points.
    }
    \label{fig:compatibility_detailed}
\end{figure*}

\subsection{Detailed Compatibility Results}
\label{app:compatibility}

Figure~\ref{fig:compatibility_detailed} further reports the per-question-type compatibility results of PlanPath with PathAgent, PathNavigate, GIANT, and BEACON. For PathAgent, PathNavigate, and GIANT, PlanPath consistently improves MCQ performance on both datasets, with gains of 2.70--5.79 percentage points on PatientVQA-Gastric and 1.65--7.02 points on PatientVQA-Renal. Larger gains are observed on Yes/No questions, reaching 17.16--27.24 points on Gastric and 1.84--7.95 points on Renal.

For open-ended questions, the effect is more method-dependent. PlanPath improves PathAgent by 5.30 and 3.05 points on Gastric and Renal, respectively, and improves PathNavigate by 1.52 and 3.06 points. In contrast, GIANT shows small decreases of 2.27 and 1.52 points on the two datasets, indicating that the benefit of patient-level orchestration can vary with the underlying agent for open-ended generation.

BEACON is evaluated separately on the question types for which outputs are available. Integrating PlanPath improves its MCQ accuracy by 10.42 points on Gastric and 4.96 points on Renal, while Yes/No accuracy increases by 17.91 and 8.25 points, respectively. Overall, these detailed results show that the improvements observed in overall accuracy are supported by consistent gains on structured question types across different underlying agents, while open-ended performance remains more dependent on the generation behavior of the base agent.


\begin{table*}[t]
\centering
\caption{
Detailed ablation results of PlanPath on PatientVQA-Gastric and PatientVQA-Renal.
Cell Quant. denotes the Cell-level Quantification Tool.
All values are reported as percentages.
}
\label{tab:ablation_detailed}

\setlength{\tabcolsep}{5pt}
\renewcommand{\arraystretch}{1.08}

\resizebox{\textwidth}{!}{
\begin{tabular}{lcccccccc}
\toprule

\multirow{2}{*}{\textbf{Configuration}}
& \multicolumn{4}{c}{\textbf{PatientVQA-Gastric}}
& \multicolumn{4}{c}{\textbf{PatientVQA-Renal}} \\

\cmidrule(lr){2-5}
\cmidrule(lr){6-9}

& \textbf{Total}
& \textbf{MCQ}
& \textbf{Yes/No}
& \textbf{Open}
& \textbf{Total}
& \textbf{MCQ}
& \textbf{Yes/No}
& \textbf{Open} \\

\midrule

\textbf{PlanPath (Full)}
& \textbf{60.24}
& \textbf{57.92}
& \textbf{70.15}
& \textbf{55.05}
& \textbf{72.71}
& \textbf{78.51}
& \textbf{74.92}
& \textbf{56.49} \\

w/o Slide Router
& 58.50
& 57.53
& 70.15
& 51.26
& 71.29
& 76.03
& 73.70
& 56.49 \\

w/o Tissue Analysis
& 56.45
& 58.69
& 64.55
& 49.49
& 69.43
& 73.97
& 72.78
& 52.67 \\

w/o Cell Quant.
& 59.59
& 58.30
& 70.15
& 53.28
& 72.57
& 78.51
& 74.92
& 55.73 \\

w/o Tissue Analysis \& Cell Quant.
& 57.31
& 56.37
& 64.18
& 53.28
& 69.57
& 73.55
& 72.78
& 54.20 \\

\bottomrule
\end{tabular}
}

\end{table*}

\subsection{Detailed Ablation Results}
\label{app:ablation_details}

Table~\ref{tab:ablation_detailed} reports the per-dataset and per-question-type ablation results corresponding to the overall results in the main text. Removing the Tissue-level Analysis Tool causes the largest and most consistent degradation, reducing Total Accuracy by 3.79 and 3.28 percentage points on PatientVQA-Gastric and PatientVQA-Renal, respectively. The Slide Router also contributes to both datasets, with particularly clear effects on Gastric open-ended questions and Renal MCQ performance. Removing the Cell-level Quantification Tool produces a smaller overall decrease, with most of its effect concentrated on open-ended questions. These results further show that the contributions of individual components are task-dependent and not strictly additive across question types.

\subsection{Detailed Quantitative Evaluation}
\label{app:quantitative}

\begin{figure*}[t]
    \centering
    \includegraphics[width=\textwidth]{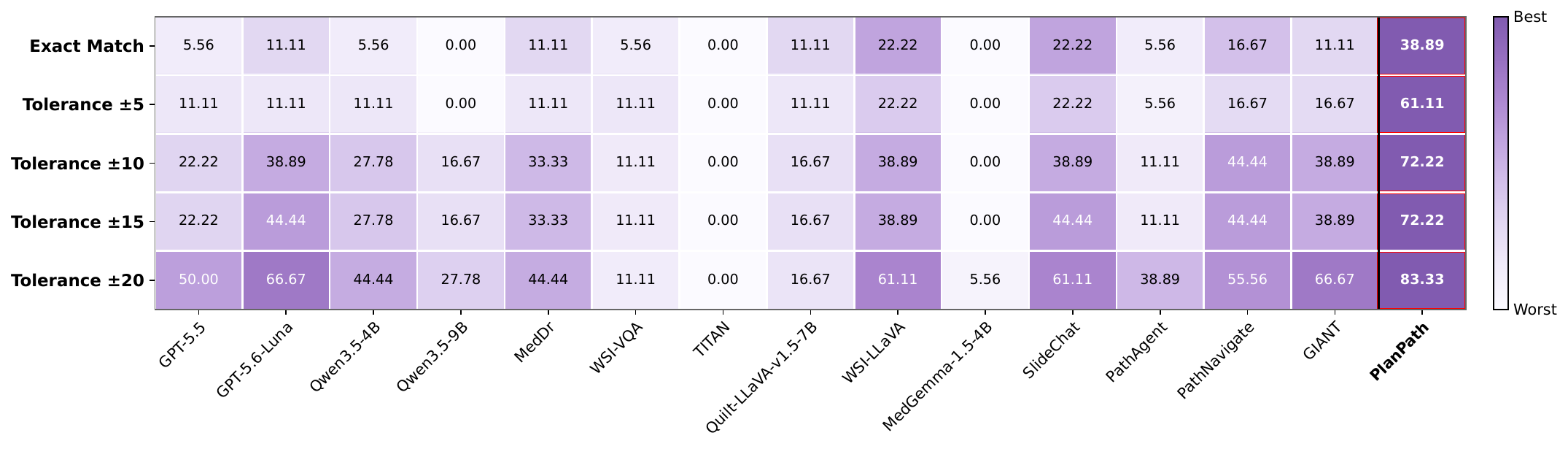}
    \caption{
    Detailed comparison of all baselines on patient-level quantitative
    questions.
    Performance is evaluated using Exact Match and tolerance-based accuracy
    with thresholds of $\pm5$, $\pm10$, $\pm15$, and $\pm20$.
    Numerical values denote accuracy (\%).
    Color intensity is normalized independently within each evaluation
    criterion, with darker purple indicating relatively better performance
    within the same row.
    Red boxes indicate the best result for each criterion.
    }
    \label{fig:quantitative_full}
\end{figure*}

In the main text, we report the quantitative-question results of representative
slide-level pathology agents, including PathAgent, PathNavigate, and GIANT, for
a compact comparison with PlanPath.
Here, we provide a more comprehensive evaluation against all baselines.
Figure~\ref{fig:quantitative_full} summarizes the Exact Match accuracy and
tolerance-based accuracy under $\pm5$, $\pm10$, $\pm15$, and $\pm20$ thresholds.

As shown in Figure~\ref{fig:quantitative_full}, PlanPath consistently achieves
the best performance across all five evaluation criteria.
Specifically, PlanPath obtains an Exact Match accuracy of 38.89\%, compared
with 22.22\% for the strongest baselines under this strict criterion.
When progressively relaxing the tolerance threshold, its accuracy increases
to 61.11\%, 72.22\%, 72.22\%, and 83.33\% under $\pm5$, $\pm10$, $\pm15$,
and $\pm20$, respectively.
The corresponding best baseline results are 22.22\%, 44.44\%, 44.44\%,
and 66.67\%.
These results show that the advantage of PlanPath is not limited to exact
numerical matching, but remains consistent across different levels of
numerical tolerance, supporting its ability to acquire and integrate
quantitative pathological evidence at the patient level.

\section{Additional Qualitative Analysis}
\label{app:qualitative}

We provide five additional examples from PatientVQA-Gastric and PatientVQA-Renal to complement the case study in Fig.~\ref{fig4}. These examples illustrate how PlanPath translates a patient-level question into an exploration plan, acquires relevant evidence across slides and magnifications, and integrates the resulting observations into a final answer. The visualizations highlight different evidence requirements for quantitative estimation, cross-slide assessment, and question-specific lesion identification.

\subsection{Visualization of Quantitative Reasoning Cases}
\label{app:quantitative_cases}

\begin{figure}[p]
\centering
\includegraphics[width=\linewidth]{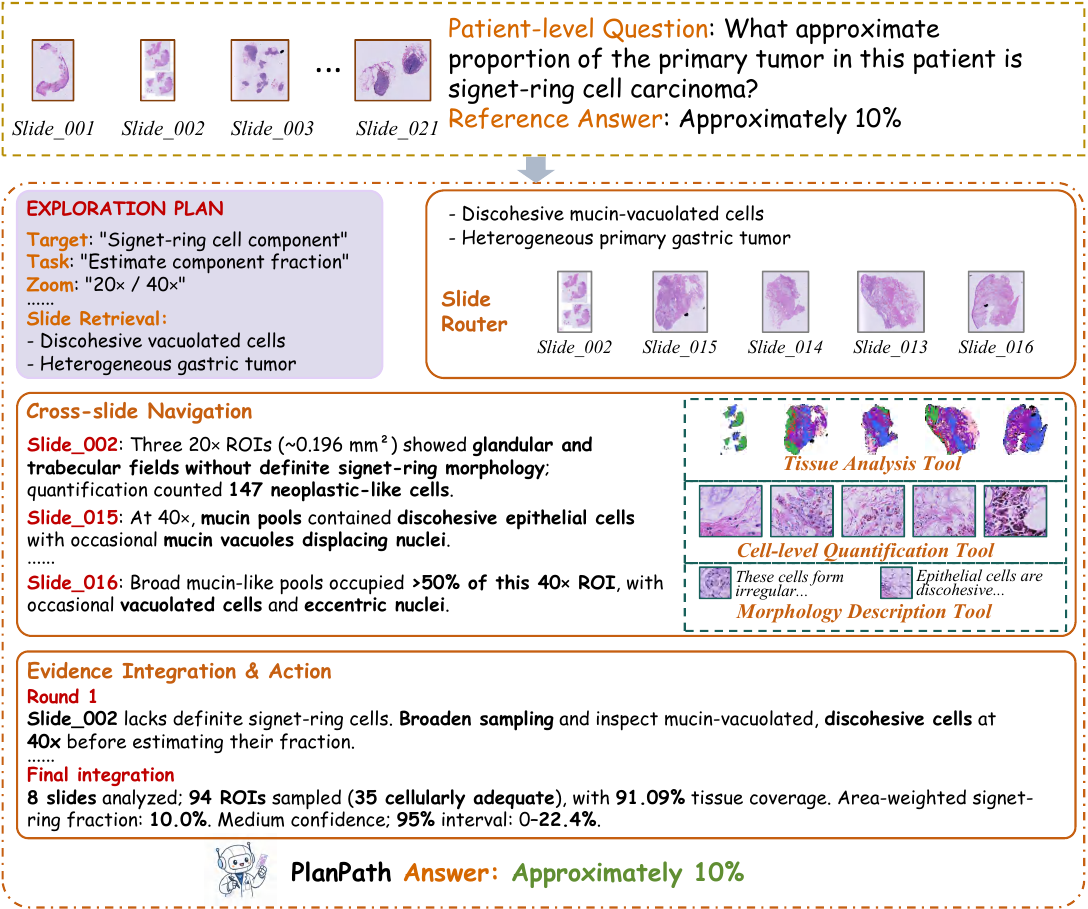}
\caption{Patient-level estimation of the signet-ring component. The exploration plan targets poorly cohesive cells with intracellular mucin and displaced nuclei. Higher-magnification observations provide phenotype-specific support, which is integrated with tissue-level quantitative evidence to estimate a proportion of approximately 10\%.}
\label{fig:app_signet10}
\end{figure}

\begin{figure}[p]
\centering
\includegraphics[width=\linewidth]{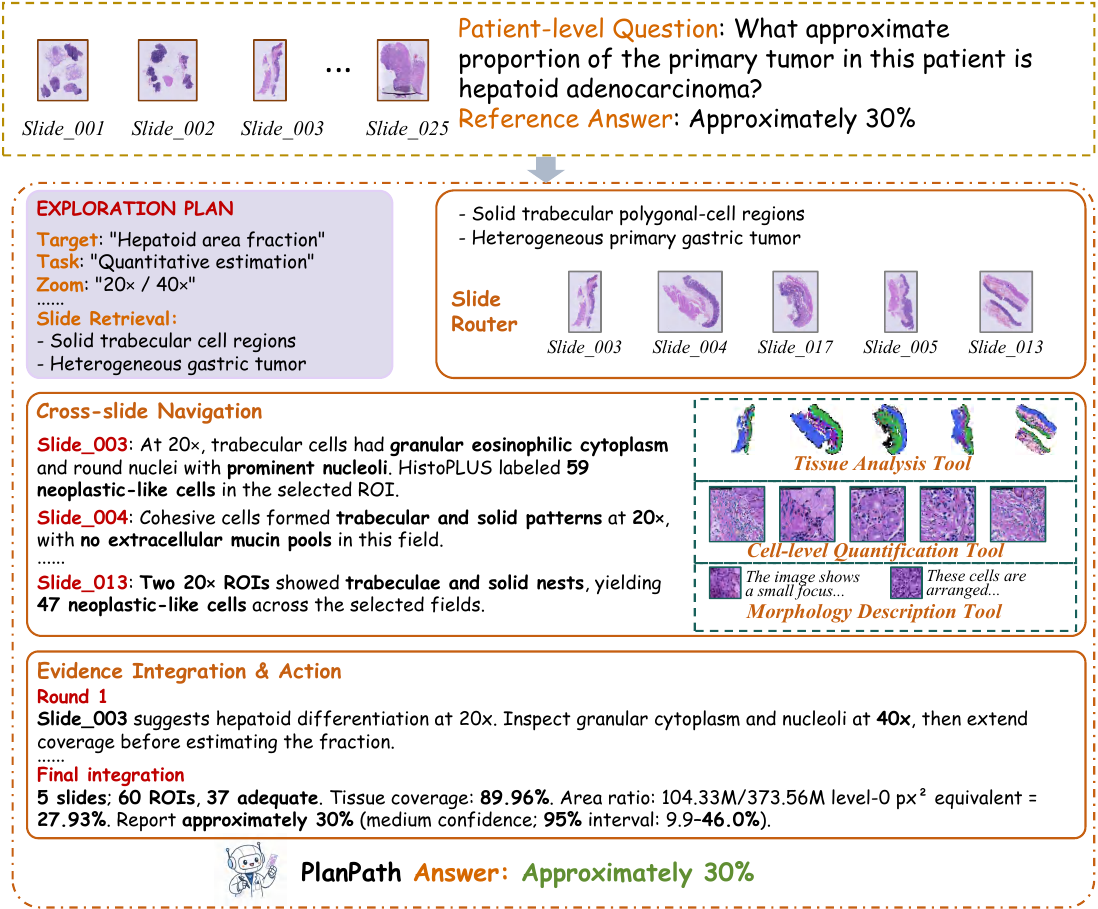}
\caption{Patient-level estimation of the hepatoid component. PlanPath examines trabecular and solid tumor regions and acquires higher-magnification evidence of polygonal cells with eosinophilic cytoplasm. Quantitative integration across the selected slides yields an estimated proportion of approximately 30\%.}
\label{fig:app_hepatoid30}
\end{figure}

Figures~\ref{fig:app_signet10} and~\ref{fig:app_hepatoid30} present two examples of histological component estimation. Answering these questions requires identifying the target morphology and assessing its extent within the tumor. PlanPath therefore combines tissue-level spatial context, ROI-level morphological observations, and cellular measurements to support area-based aggregation across the selected slides.

\textbf{Signet-ring component.}
Figure~\ref{fig:app_signet10} addresses a question targeting poorly cohesive cells with intracytoplasmic mucin vacuoles and displaced nuclei. Some 20$\times$ fields show cohesive glandular or trabecular structures without definite signet-ring morphology, prompting further exploration. Subsequent 40$\times$ observations provide supporting cellular detail, which is integrated with tissue-level information to obtain an estimate of approximately 10\%. This case illustrates how phenotype-specific retrieval and higher-magnification inspection support estimation of a component with limited extent.

\textbf{Hepatoid component.}
In Fig.~\ref{fig:app_hepatoid30}, PlanPath searches for trabecular or solid arrangements of polygonal cells. Higher-magnification observations reveal granular eosinophilic cytoplasm and prominent nucleoli in representative regions. Evidence acquired from five candidate slides supports an estimated hepatoid proportion of approximately 30\%. Together, these cases show how component-specific morphology and tissue-level extent contribute complementary information to patient-level quantitative reasoning.

\subsection{Visualization of Cross-Slide Patient-Level Reasoning}
\label{app:cross_slide_cases}

\begin{figure}[p]
\centering
\includegraphics[width=\linewidth]{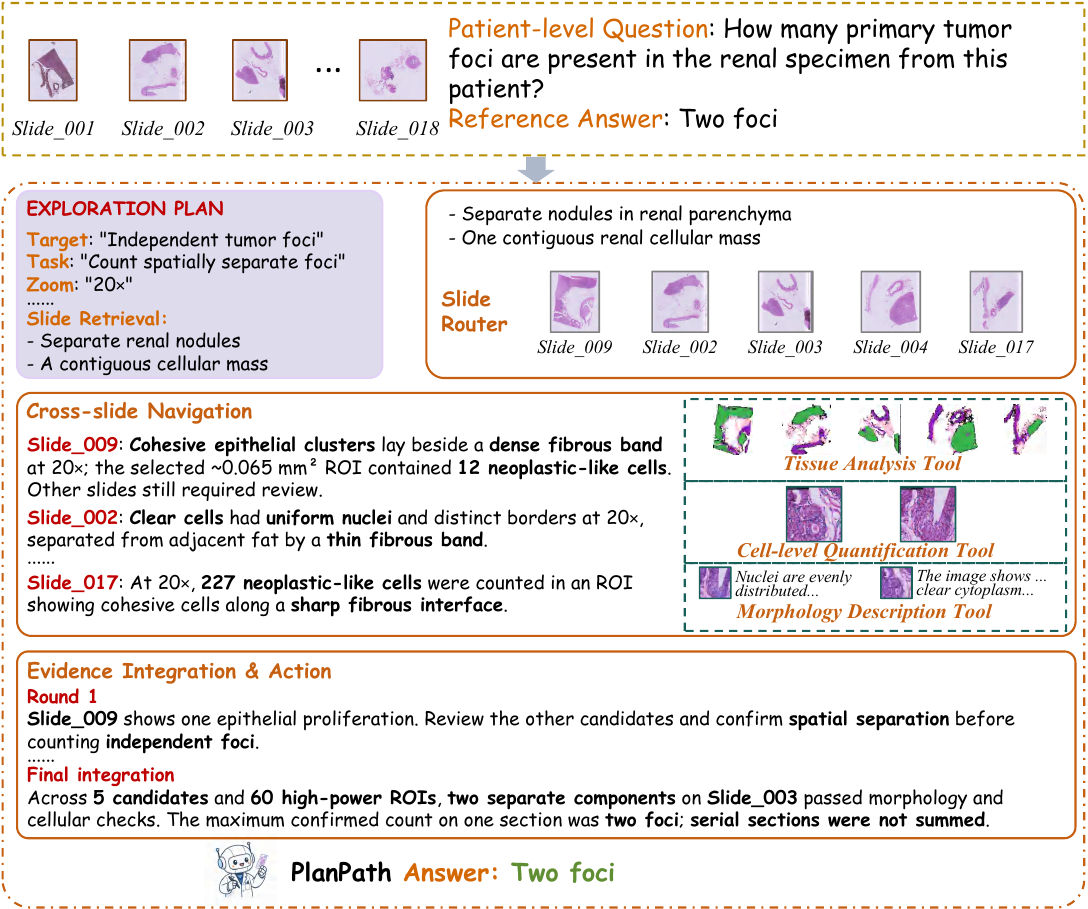}
\caption{Cross-slide assessment of primary tumor focality. Tissue-level localization identifies spatially separate candidate components, and higher-magnification observations confirm their cellular content. Integration yields two independently supported foci while avoiding repeated counting across serial sections.}
\label{fig:app_two_foci}
\end{figure}

\begin{figure}[p]
\centering
\includegraphics[width=\linewidth]{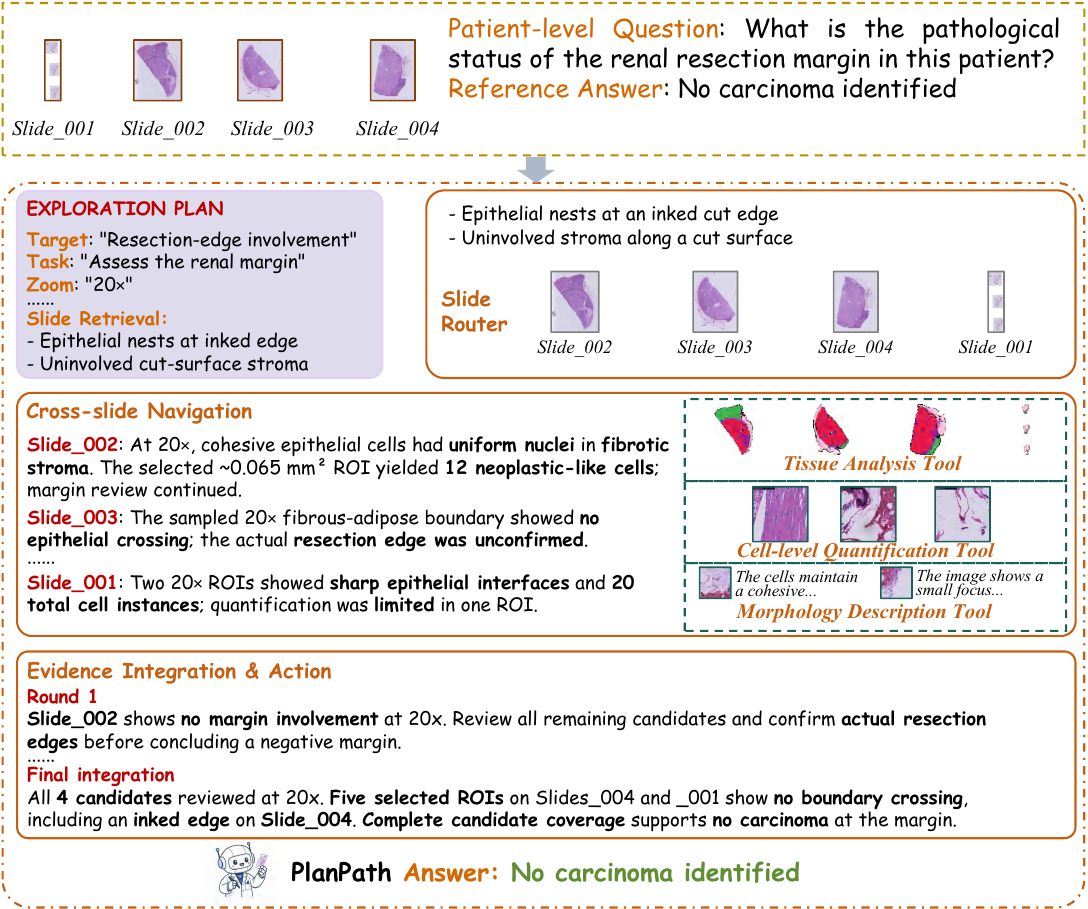}
\caption{Cross-slide assessment of the renal resection margin. PlanPath localizes relevant cut-edge relationships and continues exploration after an initial negative finding to satisfy the four-slide candidate coverage requirement. The integrated answer is no carcinoma identified at the margin.}
\label{fig:app_negative_margin}
\end{figure}

\begin{figure}[p]
\centering
\includegraphics[width=\linewidth]{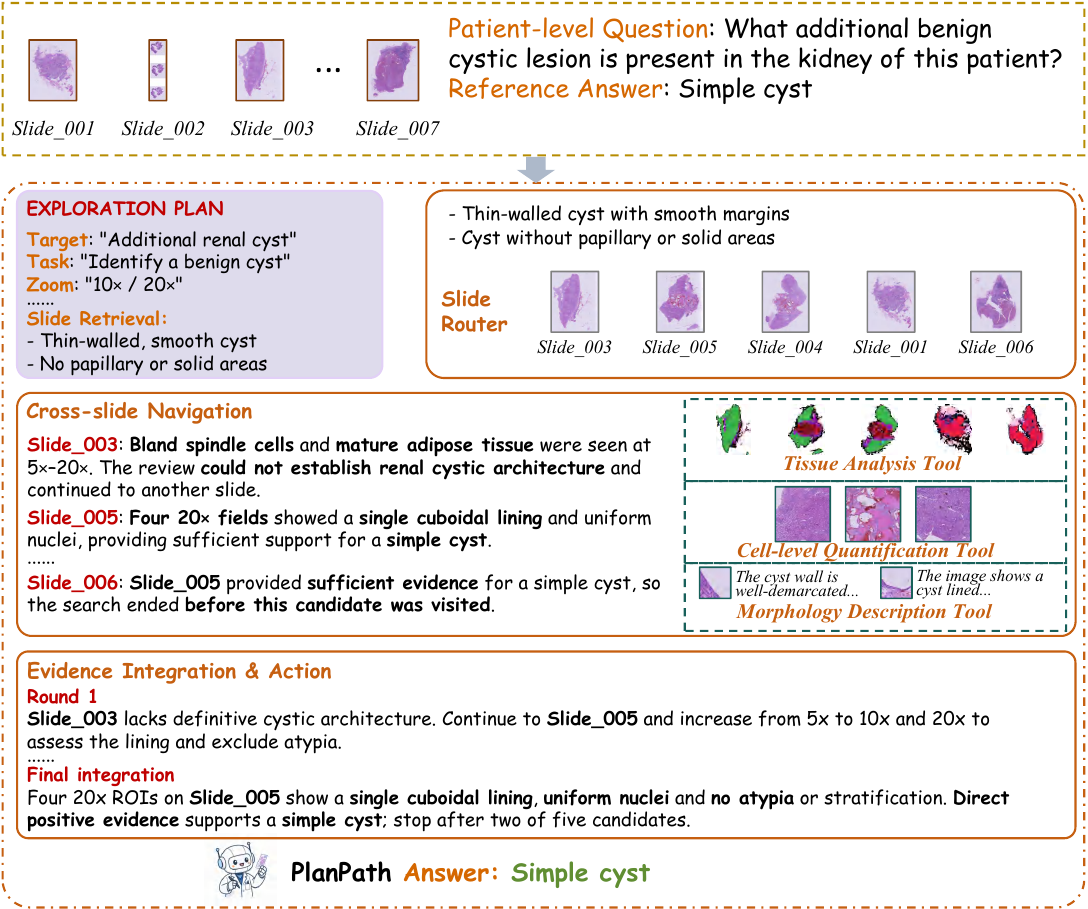}
\caption{Identification of an additional simple cyst. Targeted exploration reveals cystic architecture and a single bland cuboidal lining, providing sufficient positive evidence after two candidate slides. Cellular views provide morphological context.}
\label{fig:app_simple_cyst}
\end{figure}

Figures~\ref{fig:app_two_foci}, \ref{fig:app_negative_margin}, and~\ref{fig:app_simple_cyst} illustrate cross-slide reasoning for tumor focality, margin status, and additional lesion identification. These tasks differ in both the evidence to be acquired and the conditions under which exploration can terminate. PlanPath maintains the association between each observation and its source tissue while integrating complementary findings across the patient's slides.

\textbf{Primary tumor foci.}
Figure~\ref{fig:app_two_foci} illustrates the use of spatial relationships in focus counting. PlanPath first localizes separate tissue components and then examines their cellular content at higher magnification. After reviewing five candidate slides, it identifies two independently supported foci. Aggregation preserves the distinction between spatially separate foci and repeated appearances across sections: the final count uses the maximum independently confirmed count within a section, without summing serial-section observations.

\textbf{Renal resection margin.}
In Fig.~\ref{fig:app_negative_margin}, the exploration targets the relationship between epithelial cells and the submitted resection edge. An inked-edge observation shows no epithelial crossing, but this local finding does not immediately end the search. PlanPath continues until all four candidate slides have been examined and integrates the relevant edge and interface observations, returning \emph{no carcinoma identified}. The example illustrates the broader evidence requirement for a negative margin assessment.

\textbf{Additional simple cyst.}
In Fig.~\ref{fig:app_simple_cyst}, the first examined slide does not establish the required cystic architecture. Exploration then moves to another candidate, where 20$\times$ observations reveal a single cuboidal epithelial lining with uniform nuclei and no stratification or significant atypia. These findings support a \emph{simple cyst}, allowing PlanPath to stop after examining two of the five candidates. The decisive evidence is supplied by tissue architecture and lining-cell morphology.

\end{sloppypar}
\end{document}